%% file: continuum.tex
\documentclass{article}

\usepackage[final]{nips_2017}
\usepackage[utf8]{inputenc}
\usepackage[T1]{fontenc}
\usepackage{hyperref}
\usepackage{url}
\usepackage{booktabs}
\usepackage{amsfonts}
\usepackage{nicefrac}
\usepackage{microtype}
\usepackage{color}
\usepackage{amsmath}
\usepackage{amsthm}
\usepackage{graphicx}
\usepackage{subcaption}
\usepackage{lipsum}
\usepackage{algorithm}
\usepackage{algpseudocode}
\usepackage{xpatch}
\usepackage{fancyvrb}

\newcommand{\X}{\mathcal{X}}
\newcommand{\Y}{\mathcal{Y}}
\newcommand{\T}{\mathcal{T}}

\makeatletter
\xpatchcmd{\algorithmic}{\ALG@tlm\z@}{\ALG@tlm\z@\leftmargin 0pt}{}{}
\makeatother

\title{Gradient Episodic Memory for Continual Learning}

\author{
  David Lopez-Paz and Marc'Aurelio Ranzato\\
  Facebook Artificial Intelligence Research\\
  \texttt{\{dlp,ranzato\}@fb.com} \\
}

\begin{document}

\maketitle

\begin{abstract}
  One major obstacle towards AI is the poor ability of models to solve new
  problems quicker, and without forgetting previously acquired knowledge.  To
  better understand this issue, we study the problem of \emph{continual
  learning}, where the model observes, once and one by one, examples concerning
  a sequence of tasks.  First, we propose a set of metrics to evaluate models
  learning over a continuum of data. These metrics characterize models not only
  by their test accuracy, but also in terms of their ability to transfer
  knowledge across tasks.  Second, we propose a model for continual learning,
  called Gradient Episodic Memory (GEM) that alleviates forgetting, while
  allowing beneficial transfer of knowledge to previous tasks.  Our experiments
  on variants of the MNIST and CIFAR-100 datasets demonstrate the strong
  performance of GEM when compared to the state-of-the-art.
\end{abstract}

\section{Introduction}
The starting point in supervised learning is to collect a \emph{training set}
$D_{\text{tr}} = \{(x_i, y_i)\}_{i=1}^n$, where each \emph{example} $(x_i,
y_i)$ is composed by a \emph{feature vector} $x_i \in \X$, and a \emph{target
vector} $y_i \in \Y$. Most supervised learning methods assume that each example
$(x_i, y_i)$ is an identically and independently distributed (iid) sample from
a \emph{fixed} probability distribution $P$, which describes a single learning
task.  The goal of supervised learning is to construct a model $f : \X \to \Y$,
used to predict the target vectors $y$ associated to unseen feature vectors
$x$, where $(x, y) \sim P$. To accomplish this, supervised learning methods
often employ the {Empirical Risk Minimization} (ERM)
principle~\citep{Vapnik98}, where $f$ is found by minimizing
$\frac{1}{|D_{\text{tr}}|} \sum_{(x_i, y_i) \in D_\text{tr}} \ell(f(x_i),
y_i)$, where  $\ell : \Y \times \Y \to [0, \infty)$ is a loss function
penalizing prediction errors. In practice, ERM often requires multiple passes
over the training set. 

ERM is a major simplification from what we deem as human learning. In stark
contrast to learning machines, learning humans observe data as an ordered
sequence, seldom observe the same example twice, they can only memorize a few
pieces of data, and the sequence of examples concerns different learning tasks.
Therefore, the iid assumption, along with any hope of employing the ERM
principle, fall apart. In fact, straightforward applications of ERM lead to
``catastrophic forgetting''~\citep{catastrophic}.  That is, the learner forgets
how to solve past tasks after it is exposed to new tasks.

This paper narrows the gap between ERM and the more human-like learning
description above. In particular, our learning machine will observe,
\emph{example by example}, the \emph{continuum of data}
\begin{equation}\label{eq:continuum}
(x_1, t_1, y_1), \ldots, (x_i, t_i, y_i), \ldots, (x_n, t_n, y_n),
\end{equation}
where besides input and target vectors, the learner observes $t_i \in \T$, a
\emph{task descriptor} identifying the task associated to the pair $(x_i, y_i)
\sim P_{t_i}$.  Importantly, examples are not drawn iid from a fixed
probability distribution over triplets $(x,t,y)$, since a whole sequence of
examples from the current task may be observed before switching to the next
task.  The goal of \emph{continual learning} is to construct a model $f : \X
\times \T$ able to predict the target $y$ associated to a test pair $(x, t)$,
where $(x, y) \sim P_t$. In this setting, we face challenges unknown to ERM:
\begin{enumerate}
  \item \emph{Non-iid input data}: the continuum of data is not \emph{iid} with
  respect to any fixed probability distribution $P(X,T,Y)$ since, once tasks
  switch, a whole sequence of examples from the new task may be observed.
  \item \emph{Catastrophic forgetting}: learning new tasks may hurt the
  performance of the learner at previously solved tasks.
  \item \emph{Transfer learning}: when the tasks in the continuum are related,
  there exists an opportunity for transfer learning. This would translate into
  faster learning of new tasks, as well as performance improvements in old
  tasks.
\end{enumerate}

The rest of this paper is organized as follows. In Section~\ref{sec:continuum},
we formalize the problem of continual learning, and introduce a set of metrics
to evaluate learners in this scenario.  In Section~\ref{sec:model}, we propose
GEM, a model to learn over continuums of data that alleviates forgetting, while
transferring beneficial knowledge to past tasks.  In
Section~\ref{sec:experiments}, we compare the performance of GEM to the
state-of-the-art.  Finally, we conclude by reviewing the related literature in
Section~\ref{sec:relatedwork}, and offer some directions for future research in
Section~\ref{sec:conclusion}. Our source code is available at
\url{https://github.com/facebookresearch/GradientEpisodicMemory}.

\section{A Framework for Continual Learning}\label{sec:continuum}
We focus on the \emph{continuum} of data of
\eqref{eq:continuum}, where each triplet $(x_i, t_i, y_i)$ is formed by a
feature vector $x_i \in \X_{t_i}$, a task descriptor $t_i \in \T$, and a target
vector $y_i \in \Y_{t_i}$. For simplicity, we assume that the continuum is
\emph{locally iid}, that is, every triplet $(x_i, t_i, y_i)$ satisfies $(x_i,
y_i) \stackrel{iid}{\sim} P_{t_i}(X, Y)$.

While observing the data \eqref{eq:continuum} \emph{example by example}, our
goal is to learn a predictor $f : \X \times \T \to \Y$, which can be queried
\emph{at any time} to predict the target vector $y$ associated to a test pair
$(x, t)$, where $(x,y) \sim P_t$. Such test pair can belong to a task that we
have observed in the past, the current task, or a task that we will experience
(or not) in the future.

\paragraph{Task descriptors} An important component in our framework is the
collection of task descriptors $t_1, \ldots, t_n \in \mathcal{T}$. In the
simplest case, the task descriptors are integers $t_i = i \in \mathbb{Z}$
enumerating the different tasks appearing in the continuum of data.  More
generally, task descriptors $t_i$ could be structured objects, such as a
paragraph of natural language explaining how to solve the $i$-th task. Rich
task descriptors offer an opportunity for zero-shot learning, since the
relation between tasks could be inferred using new task descriptors alone.
Furthermore, task descriptors disambiguate similar learning tasks.  In
particular, the same input $x_i$ could appear in two different tasks, but
require different targets. Task descriptors can reference the existence of
multiple \emph{learning environments}, or provide additional (possibly
hierarchical) \emph{contextual information} about each of the examples.
However, in this paper we focus on alleviating catastrophic forgetting when
learning from a continuum of data, and leave zero-shot learning for future
research.

Next, we discuss the training protocol and evaluation metrics for continual
learning.

\subsection*{Training Protocol and Evaluation Metrics} Most of the literature
about learning over a sequence of tasks~\citep{progressive, pathnet, ewc,
icarl} describes a setting where i) the number of tasks is small, ii) the
number of examples per task is large, iii) the learner performs several passes
over the examples concerning each task, and iv) the only metric reported is the
average performance across all tasks.  In contrast, we are interested in the
``more human-like'' setting where i) the number of tasks is large, ii) the
number of training examples per task is small, iii) the learner observes the
examples concerning each task only once, and iv) we report metrics that measure
both transfer and forgetting.

Therefore, at training time we provide the learner with only one example at the
time (or a small mini-batch), in the form of a triplet $(x_i, t_i, y_i)$. The
learner never experiences the same example twice, and tasks are streamed in
sequence. We do not need to impose any order on the tasks, since a future task
may coincide with a past task.

Besides monitoring its performance across tasks, it is also important to assess
the ability of the learner to {\em transfer} knowledge. More specifically, we
would like to measure:
\begin{enumerate}
  \item \emph{Backward transfer} (BWT), which is the influence that learning a
  task $t$ has on the performance on a previous task $k \prec t$. On the one
  hand, there exists \emph{positive} backward transfer when learning about some
  task $t$ increases the performance on some preceding task $k$. On the other
  hand, there exists \emph{negative} backward transfer when learning about some
  task $t$ decreases the performance on some preceding task $k$. Large negative
  backward transfer is also known as \emph{(catastrophic) forgetting}. 
  \item \emph{Forward transfer} (FWT), which is the influence that learning a
  task $t$ has on the performance on a future task $k \succ t$. In particular,
  \emph{positive} forward transfer is possible when the model is able to
  perform ``zero-shot'' learning, perhaps by exploiting the structure available
  in the task descriptors. 
\end{enumerate}

For a principled evaluation, we consider access to a test set for each of the
$T$ tasks. After the model finishes learning about the task $t_i$, we evaluate
its \emph{test} performance on all $T$ tasks.  By doing so, we construct the
matrix $R \in \mathbb{R}^{T \times T}$, where $R_{i,j}$ is the test
classification accuracy of the model on task $t_j$ after observing the last
sample from task $t_i$.  Letting $\bar{b}$ be the vector of test accuracies for
each task at random initialization, we define three metrics:
\begin{eqnarray}
\mbox{\small {\bf Average Accuracy: } \normalsize ACC} & = & \frac{1}{T}
\sum_{i=1}^T R_{T,i} \label{eq:acc} \\
\mbox{\small {\bf Backward Transfer: } \normalsize BWT} & = & \frac{1}{T-1}
\sum_{i=1}^{T-1} R_{T,i} - R_{i,i}   \label{eq:bwt} \\
\mbox{\small {\bf Forward Transfer: } \normalsize FWT} & = & \frac{1}{T-1}
\sum_{i=2}^{T} R_{i-1,i} - \bar{b}_i. \label{eq:fwt} 
\end{eqnarray}
The larger these metrics, the better the model. If two
models have similar ACC, the most preferable one is the one with larger BWT and
FWT. Note that it is meaningless to discuss backward transfer for the first
task, or forward transfer for the last task.

For a fine-grained evaluation that accounts for learning speed, one 
can build a matrix $R$ with more rows than tasks, by evaluating more often.
In the extreme case, the number of rows could equal the number of
continuum samples $n$.  Then, the number $R_{i,j}$ is the test accuracy
on task $t_j$ after observing the $i$-th example in the continuum. Plotting
each column of $R$ results into a learning curve.

\section{Gradient of Episodic Memory (GEM)}\label{sec:model}
In this section, we propose Gradient Episodic Memory (GEM), a model for
continual learning, as introduced in Section~\ref{sec:continuum}.  The main
feature of GEM is an \emph{episodic memory} $\mathcal{M}_t$, which stores a
subset of the observed examples from task $t$.  For simplicity, we assume
integer task descriptors, and use them to index the episodic memory.  When
using integer task descriptors, one cannot expect significant positive forward
transfer (zero-shot learning). Instead, we focus on minimizing negative
backward transfer (catastrophic forgetting) by the efficient use of episodic
memory.

In practice, the learner has a total budget of $M$ memory locations.  If the
number of total tasks $T$ is known, we can allocate $m=M/T$ memories for each
task.  Conversely, if the number of total tasks $T$ is unknown, we can
gradually reduce the value of $m$ as we observe new tasks \citep{icarl}.  For
simplicity, we assume that the memory is populated with the last $m$ examples
from each task, although better memory update strategies could be employed
(such as building a coreset per task).  In the following, we consider
predictors $f_\theta$ parameterized by $\theta \in \mathbb{R}^p$, and define
the loss at the memories from the $k$-th task as
\begin{equation}
  \ell(f_\theta, \mathcal{M}_k) = \frac{1}{|\mathcal{M}_k|} \sum_{(x_i, k, y_i)
  \in \mathcal{M}_k} \ell(f_\theta(x_i, k), y_i). \label{eq:loss_past_task}
\end{equation}
Obviously, minimizing the loss at the current example together with
\eqref{eq:loss_past_task} results in overfitting to the examples stored in
$\mathcal{M}_k$.  As an alternative, we could keep the predictions at past
tasks invariant by means of distillation~\citep{icarl}.  However, this would
deem positive backward transfer impossible.  Instead, we will use the losses
\eqref{eq:loss_past_task} as {\em inequality constraints}, avoiding their
increase but allowing their decrease.  In contrast to the state-of-the-art
\citep{ewc, icarl}, our model therefore allows \textit{positive} backward
transfer.

More specifically, when observing the triplet $(x, t, y)$, we solve the
following problem:
\begin{align}
\text{minimize}_\theta\quad& \ell(f_\theta(x,t), y)\nonumber\\
\text{subject to}\quad& \ell(f_\theta, \mathcal{M}_{k}) \leq
\ell(f^{t-1}_{\theta}, \mathcal{M}_{k}) \mbox{ for all } k < t,
\label{eq:projgrad-batch}
\end{align}
where $f^{t-1}_{\theta}$ is the predictor state at the end of learning of task
$t-1$.

In the following, we make two key observations to
solve~\eqref{eq:projgrad-batch} efficiently.  First, it is unnecessary to store
old predictors $f^{t-1}_\theta$, as long as we guarantee that the loss at
previous tasks does not increase after each parameter update $g$.  Second,
assuming that the function is locally linear (as it happens around small
optimization steps) and that the memory is representative of the examples from
past tasks, we can diagnose increases in the loss of previous tasks by
computing the angle between their loss gradient vector and the proposed update.
Mathematically, we rephrase the constraints~\eqref{eq:projgrad-batch} as:
\begin{equation}
\left\langle g, g_{k} \right\rangle := 
\left
\langle
\frac{\partial \ell(f_\theta(x, t), y)}
{\partial \theta}, 
\frac{\partial \ell(f_\theta, \mathcal{M}_k)}
{\partial \theta}
\right\rangle \ge 0, \mbox{ for all } 
 k < t.
\label{eq:constraints}
\end{equation}
If all the inequality constraints~\eqref{eq:constraints} are satisfied, then
the proposed parameter update $g$ is unlikely to increase the loss at previous
tasks.  On the other hand, if one or more of the inequality
constraints~\eqref{eq:constraints} are violated, then there is at least one
previous task that would experience an increase in loss after the parameter
update.  If violations occur, we propose to project the proposed gradient $g$
to the closest gradient $\tilde{g}$ (in squared $\ell_2$ norm) satisfying all
the constraints~\eqref{eq:constraints}. Therefore, we are interested in: 
\begin{align}
\text{minimize}_{\tilde{g}} \frac{1}{2} \quad& \|g - \tilde{g}\|_2^2\nonumber\\
\text{subject to} \quad& \langle \tilde{g}, g_k \rangle \geq 0 \text{ for all } k < t.\label{eq:gemprimal}
\end{align}

To solve \eqref{eq:gemprimal} efficiently, recall the primal of a Quadratic
Program (QP) with inequality constraints:
\begin{align}
    \text{minimize}_z \quad&\frac{1}{2} z^\top C z + p^\top z\nonumber\\
    \text{subject to} \quad&Az \geq b,\label{eq:primal}
\end{align}
where $C \in \mathbb{R}^{p \times p}$, $p\in \mathbb{R}^p$, $A \in
\mathbb{R}^{(t-1) \times p}$, and $b\in\mathbb{R}^{t-1}$.  The dual problem of~\eqref{eq:primal} 
is:
\begin{align}
    \text{minimize}_{u,v}   \quad&\frac{1}{2} u^\top C u - b^\top v\nonumber\\
    \text{subject to}       \quad&A^\top v - Cu = p,\nonumber\\
                                 &v \geq 0.\label{eq:dual}
\end{align}
If $(u^\star, v^\star)$ is a solution to \eqref{eq:dual}, then there is a
solution $z^\star$ to \eqref{eq:primal} satisfying $Cz^\star = Cu^\star$
\citep{dorn1960duality}.
Quadratic programs are at the heart of support vector
machines \citep{scholkopf2001learning}.

With these notations in hand, we write the primal GEM QP \eqref{eq:gemprimal} as:
\begin{align*}
    \text{minimize}_z   \quad& \frac{1}{2} z^\top z -g^\top z + \frac{1}{2} g^\top g\\
    \text{subject to}   \quad& Gz \geq 0,
\end{align*}
where $G = (g_1, \ldots, g_{t-1})$, and we discard the constant term $g^\top g$.
This is a QP on $p$ variables (the number of parameters of the neural
network), which could be measured in the millions. However, we can pose the
dual of the GEM QP as: 
\begin{align}
    \text{minimize}_{v}   \quad&\frac{1}{2} v^\top GG^\top v + g^\top G^\top v\nonumber\\
    \text{subject to} \quad&v \geq 0,\label{eq:project}
\end{align}
since $u = G^\top v + g$ and the term $g^\top g$ is constant.  This is a QP on
$t-1 \ll p$ variables, the number of observed tasks so far. Once we solve the
dual problem \eqref{eq:project} for $v^\star$, we can recover the projected
gradient update as $\tilde{g} = G^\top v^\star  + g$.  In practice, we found
that adding a small constant $\gamma \geq 0$ to $v^\star$ biased the gradient
projection to updates that favoured benefitial backwards transfer.

Algorithm~\ref{alg:gem:train} summarizes the training and evaluation protocol
of GEM over a continuum of data. The pseudo-code includes the computation of
the matrix R, containing the sufficient statistics to compute the metrics ACC,
FWT, and BWT described in Section~\ref{sec:continuum}.

\paragraph{A causal compression view} We can interpret GEM as a model that
learns the subset of correlations common to a set of distributions (tasks).
Furthermore, GEM can (and will in our MNIST experiments) be used to predict
target vectors associated to previous or new tasks without making use of task
descriptors. This is a desired feature in causal inference problems, since
causal predictions are invariant across different environments \citep{icp}, and
therefore provide the most compressed representation of a set of distributions
\citep{causal_compression}.
 
\begin{algorithm}
    \caption{Training a GEM over an \emph{ordered} continuum of data}
    \label{alg:gem:train}
    \begin{minipage}[t]{.5\linewidth}
    \begin{algorithmic}
    \small
    \Procedure{Train}{$f_\theta, \mbox{Continuum}_{\text{train}}, \mbox{Continuum}_{\text{test}}$} 
        \State $\mathcal{M}_t \leftarrow \lbrace\rbrace$ for all $t = 1, \ldots, T$.
        \State $R \leftarrow 0 \in \mathbb{R}^{T\times T}$. 
        \For{$t = 1, \ldots, T$}:
        \For{$(x,y)$ in $\mbox{Continuum}_{\text{train}}(t)$}
            \State $\mathcal{M}_t \leftarrow \mathcal{M}_t \cup (x,y)$
            \State $g \leftarrow \nabla_\theta \, \ell (f_\theta(x, t), y)$
            \State $g_{k} \leftarrow \nabla_\theta \, \ell (f_\theta, \mathcal{M}_{k})$ for all
                $k < t$ 
            \State $\tilde{g} \leftarrow$ \textsc{Project}($g, g_1, \ldots, g_{t-1}$), see \eqref{eq:project}.
            \State $\theta \leftarrow \theta - \alpha \tilde{g}$.
        \EndFor
        \State $R_{t,:} \leftarrow \textsc{Evaluate}(f_\theta, \text{Continuum}_{\text{test}})$
        \EndFor
        \State \textbf{return} $f_\theta$, R
    \EndProcedure
    \end{algorithmic}
    \end{minipage}
    \hfill
    \begin{minipage}[t]{.4\linewidth}
    \begin{algorithmic}
    \small
    \Procedure{Evaluate}{$f_\theta, \mbox{Continuum}$} 
        \State $r \leftarrow 0 \in \mathbb{R}^T$
        \For{$k = 1, \ldots, T$}
        \State $r_k \leftarrow 0$
        \For{$(x,y)$ in $\mbox{Continuum}(k)$} 
            \State $r_k \leftarrow r_k +  \mbox{accuracy}(f_\theta(x, k), y)$
        \EndFor
        \State $r_k \leftarrow r_k \,/\, \text{len}(\text{Continuum}(k))$
        \EndFor
        \State \textbf{return} $r$
    \EndProcedure
    \end{algorithmic}
    \end{minipage}
\end{algorithm}

\section{Experiments}\label{sec:experiments}
We perform a variety of experiments to assess the
performance of GEM in continual learning.

\subsection{Datasets}
We consider the following datasets:
\begin{itemize}
  \item {\em MNIST Permutations}~\citep{ewc}, a variant of the MNIST
  dataset of handwritten digits~\citep{mnist}, where each task is transformed 
  by a fixed permutation of pixels. In this dataset, the input distribution
  for each task is unrelated.
  \item {\em MNIST Rotations}, a variant of MNIST where each task contains digits
  rotated by a fixed angle between $0$ and $180$ degrees.
  \item {\em Incremental CIFAR100}~\citep{icarl}, a variant of the CIFAR object
  recognition dataset with 100 classes~\citep{cifar100}, where each task
  introduces a new set of classes. For a total number of $T$ tasks, each new
  task concerns examples from a disjoint subset of $100/T$ classes. Here, the
  input distribution is similar for all tasks, but different tasks require 
  different output distributions.
\end{itemize}

For all the datasets, we considered $T=20$ tasks. On the MNIST datasets, each
task has 1000 examples from 10 different classes. On the CIFAR100 dataset
each task has 2500 examples from 5 different classes.  The model observes the
tasks in sequence, and each example once. The evaluation for each task is
performed on the test partition of each dataset. 

\subsection{Architectures}
On the MNIST tasks, we use fully-connected neural networks with two hidden
layers of $100$ ReLU units.  On the CIFAR100 tasks, we use a smaller version of
ResNet18~\citep{resnet}, with three times less feature maps across all layers.
Also on CIFAR100, the network has a final linear classifier per task.
This is one simple way to leverage the task descriptor, in order to adapt the
output distribution to the subset of classes for each task.  We train all the
networks and baselines using plain SGD on mini-batches of $10$ samples. All
hyper-parameters are optimized using a grid-search (see
Appendix~A), and the best results for each model are reported.

\subsection{Methods}
We compare GEM to five alternatives:
\begin{enumerate}
  \item a {\em single} predictor trained across all tasks.
  \item one \emph{independent} predictor per task. Each \emph{independent}
  predictor has the same architecture as ``single'' but with $T$ times less
  hidden units than ``single''. Each new independent predictor can be
  initialized at   random, or be a clone of the last trained predictor (decided
  by grid-search).
  \item a \emph{multimodal} predictor, which has the same architecture of
  ``single'', but with a dedicated input layer per task (only for MNIST
  datasets). 
  \item \emph{EWC}~\citep{ewc}, where the loss is regularized to avoid
  catastrophic forgetting.
  \item \emph{iCARL}~\citep{icarl}, a class-incremental learner that classifies
  using a nearest-exemplar algorithm, and prevents catastrophic forgetting by
  using an episodic memory. iCARL requires the same input representation across
  tasks, so this method only applies to our experiment on CIFAR100. 
\end{enumerate}
GEM, iCaRL and EWC have the same architecture as ``single'', plus episodic memory. 

\subsection{Results}
\begin{figure}
  \begin{center}
  \includegraphics[width=0.49\textwidth]{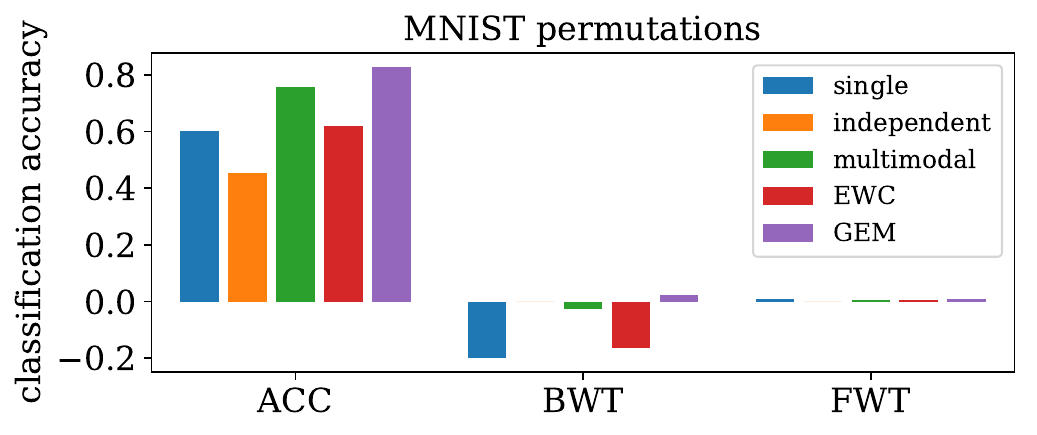}\hfill\includegraphics[width=0.49\textwidth]{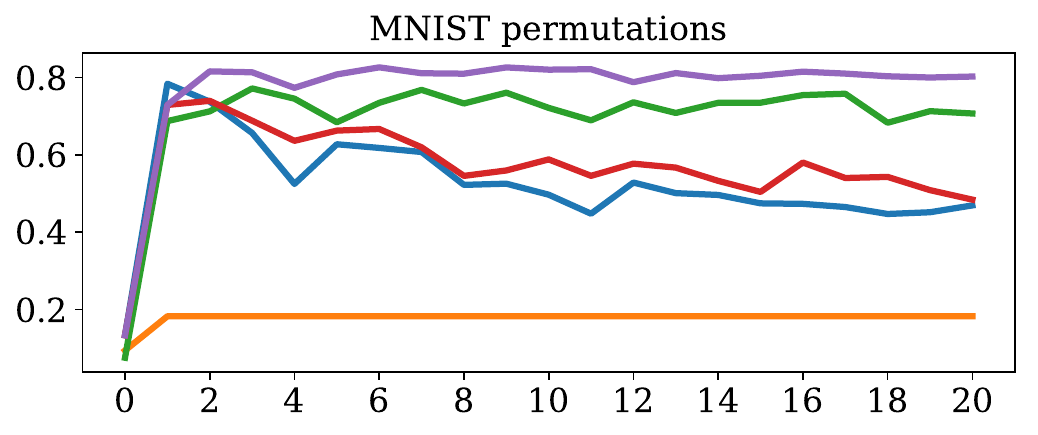}
  \vskip 0.5cm
 \includegraphics[width=0.49\textwidth]{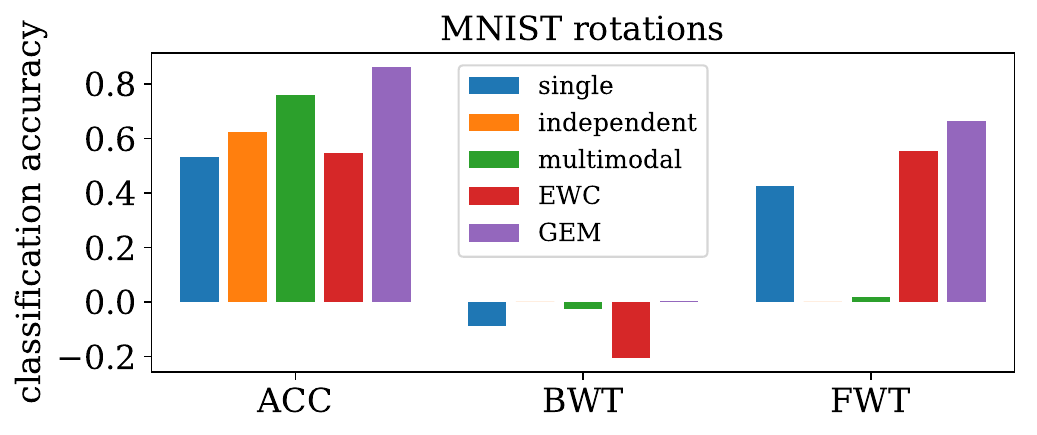}\hfill\includegraphics[width=0.49\textwidth]{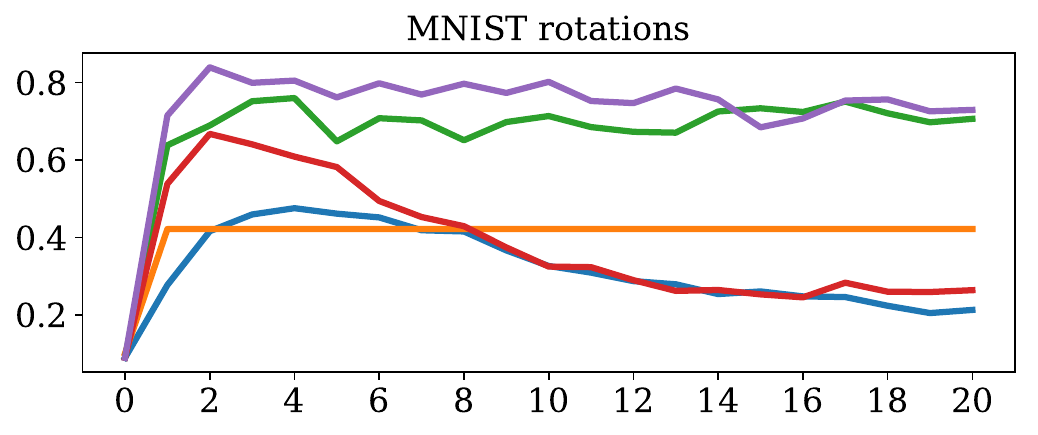}
  \vskip 0.5cm
  \includegraphics[width=0.49\textwidth]{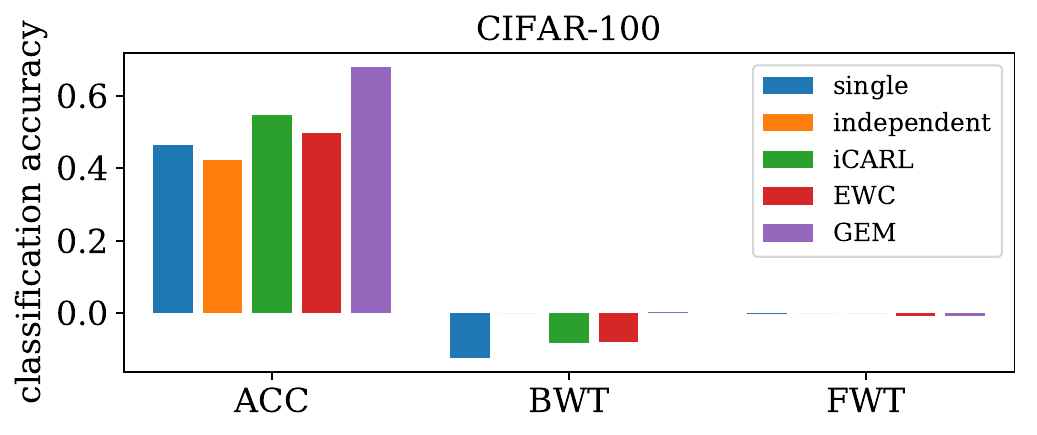}\hfill\includegraphics[width=0.49\textwidth]{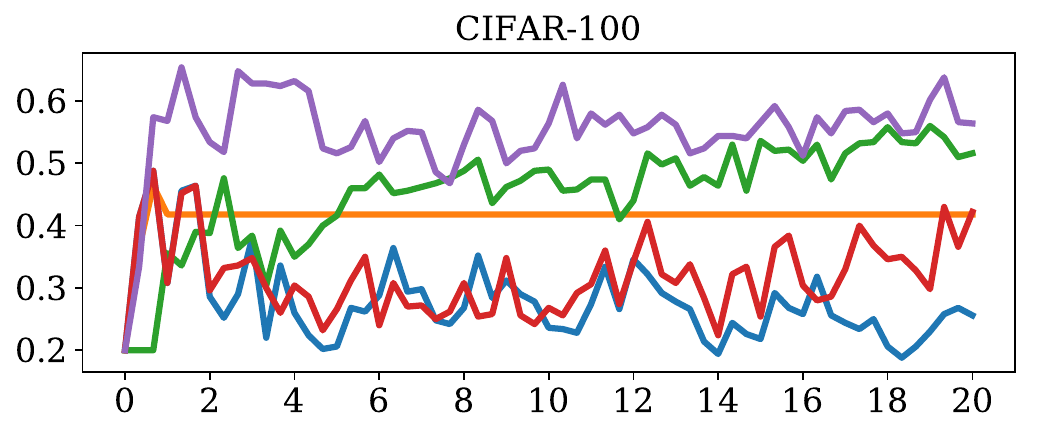}
  \end{center}
  \caption{Left: ACC, BWT, and FWT for all datasets and methods.
  Right: evolution of the test accuracy at the first task, as more tasks are learned.}
  \label{figure:results}
\end{figure}
\begin{table}
  \caption{CPU Training time (s) of MNIST experiments for all methods.}
  \begin{center}
  \begin{tabular}{lrrrrr}
  \toprule
  task & single & independent & multimodal & EWC & GEM\\
  \midrule
  permutations & $11$ & $11$ & $14$  & $179$  & $77$\\
  rotations    & $11$ & $16$ & $13$  & $169$ & $135$\\
  \bottomrule
  \end{tabular}
  \end{center}
  \label{table:timing}
\end{table}

Figure~\ref{figure:results} (left) summarizes the average accuracy (ACC,
Equation~\ref{eq:acc}), backward transfer (BWT, Equation~\ref{eq:bwt}) and
forward transfer (FWT, Equation~\ref{eq:fwt}) for all datasets and methods.  We
provide the full evaluation matrices $R$ in Appendix~B.
Overall, GEM performs similarly or better than the multimodal model (which is
very well suited to the MNIST tasks). GEM minimizes backward transfer, while
exhibiting negligible or positive forward transfer.

Figure~\ref{figure:results} (right) shows the evolution of the test accuracy of
the first task throughout the continuum of data. GEM exhibits minimal 
forgetting, and positive backward transfer in
CIFAR100. 

Overall, GEM performs significantly better than other continual learning
methods like EWC, while spending less computation (Table~\ref{table:timing}).
GEM's efficiency comes from optimizing over a number of variables equal to the
number of tasks ($T=20$ in our experiments), instead of optimizing over a
number of variables equal to the number of parameters ($p=1109240$ for CIFAR100
for instance). GEM's bottleneck is the necessity of computing previous task
gradients at each learning iteration.

\subsubsection{Importance of memory, number of passes, and order of tasks}
Table~\ref{table:memory} shows the final ACC in the CIFAR-100 experiment for
both GEM and iCARL as a function their episodic memory size. Also seen in
Table~\ref{table:memory}, the final ACC of GEM is an increasing function of the
size of the episodic memory, eliminating the need to carefully tune this
hyper-parameter. GEM outperforms iCARL for a wide range of memory sizes.
\begin{table}
  \caption{ACC as a function of the episodic memory size for GEM and iCARL, on
  CIFAR100.}
  \begin{center}
  \begin{tabular}{lrrrr}
  \toprule
  memory size & $200$ & $1,280$ & $2,560$ & $5,120$\\
  \midrule
  GEM & $0.487$ & $0.579$ & $0.633$ & $0.654$ \\
  iCARL & $0.436$ & $0.494$ & $0.500$  & $0.508$ \\
  \bottomrule
  \end{tabular}
  \end{center}
  \label{table:memory}
\end{table}

Table~\ref{table:varying_epochs} illustrates the importance of memory as we do
more than one pass through the data on the MNIST rotations experiment. Multiple
training passe exacerbate the catastrophic forgetting problem.  For instance,
in the last column of Table~\ref{table:varying_epochs} (except for the result
in the first row), each model is shown examples of a task five times (in random
order) before switching to the next task.  Table~\ref{table:varying_epochs}
shows that memory-less methods (like ``single'' and ``multimodal'') exhibit
higher negative BWT, leading to lower ACC.  On the other hand, memory-based
methods such as EWC and GEM lead to higher ACC  as the number of passes through
the data increases.  However, GEM suffers less negative BWT than EWC, leading
to a higher ACC.

Finally, to relate the performance of GEM to the best possible performance on
the proposed datasets, the first row of Table~\ref{table:varying_epochs}
reports the ACC of ``single'' when trained with iid data from all tasks. This
mimics usual multi-task learning, where each mini-batch contains examples taken
from a random selection of tasks.  By comparing the first and last row of
Table~\ref{table:varying_epochs}, we see that GEM matches the ``oracle
performance upper-bound'' ACC provided by iid learning, and minimizes negative
BWT.

\begin{table}
  \caption{ACC/BWT on the MNIST Rotations dataset, when varying the number of
  epochs per task.}
  \begin{center}
  \begin{tabular}{lrrr}
  \toprule
  method & 1 epoch & 2 epochs & 5 epochs \\
  \midrule
  single, shuffled data  &  0.83/{\color{white}-0.00} & 0.87/{\color{white}-0.00} & 0.89/{\color{white}-0.00} \\
  single      & 0.53/-0.08 & 0.49/-0.25 & 0.43/-0.40 \\
  independent & 0.56/-0.00 & 0.64/-0.00 & 0.67/-0.00 \\
  multimodal  & 0.76/-0.02 & 0.72/-0.11 & 0.59/-0.28 \\
  EWC         & 0.55/-0.19 & 0.59/-0.17 & 0.61/-0.11 \\
  GEM         & 0.86/+0.05 & 0.88/+0.02 & 0.89/-0.02 \\
  \bottomrule
  \end{tabular}
  \end{center}
  \label{table:varying_epochs}
\end{table}

\section{Related work}\label{sec:relatedwork}
Continual learning~\citep{ring1994}, also called \emph{lifelong
learning}~\citep{thrun-iros94, learning_to_learn, lifelong, nthing}, considers
learning through a sequence of tasks, where the learner has to retain knowledge
about past tasks and leverage that knowledge to quickly acquire new skills.
This learning setting led to implementations~\citep{nell, ella, child}, and
theoretical investigations~\citep{baxter, balcan_lifelong,pentina16}, although
the latter ones have been restricted to {linear} models. In this work, we
revisited continual learning but proposed to focus on the more realistic
setting where examples are seen only once, memory is finite, and the learner is
also provided with (potentially structured) task descriptors.  Within this
framework, we introduced a new set of metrics, a training and testing protocol,
and a new algorithm, GEM, that outperforms the current state-of-the-art in
terms of limiting forgetting.

The use of task descriptors is similar in spirit to recent work in
Reinforcement Learning~\citep{horde, uvfa}, where task or goal descriptors are
also fed as input to the system.  The \emph{CommAI project} \citep{commai,
commai2} shares our same motivations, but focuses on highly structured task
descriptors, such as strings of text. In contrast, we focus on the problem of
catastrophic forgetting~\citep{catastrophic, catastrophic_french,
connectionist, hippocampus, catastrophic_empirical}.

Several approaches have been proposed to avoid catastrophic forgetting.  The
simplest approach in neural networks is to freeze early layers, while cloning
and fine-tuning later layers on the new task~\citep{oquab} (which we considered
in our ``independent'' baseline). This relates to methods that leverage a
modular structure of the network with primitives that can be shared across
tasks~\citep{progressive, pathnet, expert_gate, denoyer15, eigen14}.
Unfortunately, it has been very hard to scale up these methods to lots of
modules and tasks, given the combinatorial number of compositions of modules. 

Our approach is most similar to the regularization approaches that     consider
a single model, but modify its learning objective to prevent catastrophic
forgetting.  Within this class of methods, there are approaches that leverage
``synaptic'' memory \citep{ewc, synaptic}, where learning rates are adjusted to
minimize changes in parameters important for previous tasks.  Other approaches
are instead based on ``episodic'' memory \citep{less_forgetting, lwc,
encoder_lifelong, icarl}, where examples from previous tasks are stored and
replayed to maintain predictions invariant by means of distillation
\citep{distillation}. GEM is related to these latter approaches but, unlike
them, allows for positive backward transfer. 

More generally, there are a variety of setups in the machine learning
literature related to continual learning.  \emph{Multitask learning}
\citep{multitask} considers the problem of maximizing the performance of a
learning machine across a variety of tasks, but the setup assumes simultaneous
access to all the tasks at once. Similarly, \emph{transfer learning}
\citep{transfer_learning} and \emph{domain adaptation}
\citep{domain_adaptation} assume the simultaneous availability of multiple
learning tasks, but focus at improving the performance at one of them in
particular.  \emph{Zero-shot learning} \citep{zero_lampert, zero_palatucci} and
\emph{one-shot learning} \citep{one_shot, matchingnet, santoro16, learnet} aim
at performing well on unseen tasks, but ignore the catastrophic forgetting of
previously learned tasks.  \emph{Curriculum learning} considers learning a
sequence of data \citep{curriculum}, or a sequence of tasks
\citep{curriculum_tasks}, sorted by increasing difficulty. 

\section{Conclusion}\label{sec:conclusion}
We formalized the scenario of \emph{continual learning}. First, we defined
training and evaluation protocols to assess the quality of models in terms of
their \emph{accuracy}, as well as their ability to transfer knowledge {\em
forward} and {\em backward} between tasks.  Second, we introduced GEM, a simple
model that leverages an episodic memory to avoid forgetting and favor positive
backward transfer. Our experiments demonstrate the competitive performance of
GEM against the state-of-the-art.

GEM has three points for improvement. First, GEM does not leverage structured
task descriptors, which may be exploited to obtain positive forward transfer
(zero-shot learning). Second, we did not investigate advanced memory management
(such as building \emph{coresets of tasks}~\citep{coresetMoG}). Third, each GEM
iteration requires one backward pass per task, increasing computation time.
These are exciting research directions to extend learning machines beyond ERM,
and to continuums of data.

\clearpage
\newpage

\section*{Acknowledgements}
We are grateful to M. Baroni, L. Bottou, M. Nickel, Y. Olivier and A. Szlam for
their insight. We are grateful to Martin Arjovsky for the QP interpretation of
GEM.

\bibliographystyle{abbrvnat}
{
    \small
    \bibliography{continuum}
}

\clearpage
\newpage
\appendix
\input{appendix}

\end{document}

%% file: appendix.tex
% \documentclass{article}
% 
% \usepackage[final]{nips_2017}
% \usepackage[utf8]{inputenc}
% \usepackage{microtype}
% \usepackage{fancyvrb}
% \usepackage{graphicx}
% 
% \newtheorem{proposition}{Proposition}
% 
% \title{Supplementary material for:\\ \emph{Gradient Episodic Memory for Continual Learning}}
% 
% 
% \author{
%   David Lopez-Paz and Marc'Aurelio Ranzato\\
%   Facebook Artificial Intelligence Research\\
%   \texttt{\{dlp,ranzato\}@fb.com} \\
% }
% 
% \begin{document}
% \maketitle
% \appendix

\section{Hyper-parameter Selection} \label{sec:apphyper}
Here we report the hyper-parameter grids considered in our experiments. The
best values for the MNIST rotations (rot), MNIST permutations (perm) and
CIFAR-100 incremental (cifar) experiments are noted accordingly in parenthesis.
For more details, please refer to our implementation, linked in the main text.
\begin{itemize}
  \item \emph{single}
  \begin{itemize}
    \item learning rate: [0.001, 0.003 (rot), 0.01, 0.03 (perm), 0.1, 0.3, 1.0 (cifar)]
  \end{itemize}
  \item \emph{independent}
  \begin{itemize}
    \item learning rate: [0.001, 0.003, 0.01, 0.03 (perm), 0.1 (rot), 0.3 (cifar), 1.0]
    \item finetune: [no, yes (rot, perm, cifar)]
  \end{itemize}
  \item \emph{multimodal}
  \begin{itemize}
    \item learning rate: [0.001, 0.003, 0.01, 0.03, 0.1 (rot, perm), 0.3, 1.0]
  \end{itemize}
  \item \emph{EWC}
  \begin{itemize}
    \item learning rate: [0.001, 0.003, 0.01 (rot), 0.03, 0.1 (perm), 0.3, 1.0 (cifar)]
    \item regularization: [1 (cifar), 3 (perm), 10, 30, 100, 300, 1000 (rot), 3000, 10000, 30000]
  \end{itemize}
  \item \emph{iCARL}
  \begin{itemize}
    \item learning rate: [0.001, 0.003, 0.01, 0.03, 0.1, 0.3, 1.0 (cifar)]
    \item regularization: [0.1, 0.3, 1 (cifar), 3, 10, 30]
    \item memory size:    [200, 1280, 2560, 5120 (cifar)] 
  \end{itemize}
  \item \emph{GEM}
  \begin{itemize}
    \item learning rate: [0.001, 0.003, 0.01, 0.03, 0.1 (rot, perm, cifar), 0.3, 1.0]
    \item memory size: [5120 (rot, perm, cifar)]
    \item $\gamma$: [0.0, 0.1, ..., 0.5 (rot, perm, cifar), ..., 1.0]
  \end{itemize}
\end{itemize}

\clearpage
\newpage
\section{Full experiments} \label{sec:app-r}
In this section we report the evaluation matrices $R$ for each model and
dataset.  The first row of each matrix (above the line) is the baseline test
accuracy $\bar{b}$ before training starts. The rest of entries $(i,j)$ of the
matrix $R$ report the test accuracy of the $j$-th task just after finishing
training the $i$-th task.

\subsection{MNIST permutations}
\subsubsection{Model \emph{single}}
{\tiny
\begin{Verbatim}
0.1330 0.1199 0.1070 0.0825 0.0609 0.0832 0.1385 0.1123 0.0736 0.1190 0.0666 0.0890 0.0885 0.0723 0.1083 0.0524 0.0976 0.0871 0.1143 0.0743
|
0.7838 0.1264 0.0709 0.1142 0.0671 0.0928 0.0925 0.0878 0.0759 0.0832 0.0748 0.1057 0.0927 0.0770 0.0886 0.0995 0.1328 0.0839 0.1802 0.0609
0.7374 0.7939 0.0981 0.1496 0.0859 0.0626 0.1226 0.0666 0.1045 0.0596 0.0787 0.0938 0.0948 0.0737 0.0797 0.0670 0.1450 0.1062 0.1438 0.0714
0.6564 0.7196 0.7783 0.1177 0.0603 0.0908 0.1296 0.0654 0.1246 0.0894 0.0721 0.0798 0.0811 0.0631 0.0759 0.0726 0.1338 0.1265 0.1567 0.1039
0.5246 0.6359 0.7593 0.7790 0.0612 0.0864 0.1118 0.0815 0.1028 0.0900 0.0991 0.0773 0.0947 0.0948 0.0931 0.0930 0.1111 0.1027 0.1504 0.1003
0.6272 0.7265 0.7737 0.7695 0.7827 0.0981 0.0898 0.0530 0.1031 0.0760 0.0675 0.0569 0.0989 0.0698 0.0810 0.0483 0.1305 0.0819 0.1386 0.0961
0.6177 0.7317 0.7890 0.7860 0.7808 0.8135 0.1020 0.0668 0.1005 0.0648 0.0456 0.0705 0.0932 0.0858 0.1056 0.0797 0.1483 0.1039 0.1442 0.0788
0.6071 0.6620 0.7623 0.7540 0.7657 0.7838 0.8113 0.0991 0.1127 0.0957 0.0493 0.0916 0.0713 0.0804 0.1383 0.0639 0.1336 0.1192 0.1336 0.0921
0.5221 0.6675 0.7466 0.6990 0.7239 0.7707 0.8156 0.7909 0.1107 0.0858 0.0510 0.0764 0.0668 0.0849 0.1103 0.0441 0.1232 0.0996 0.1434 0.0744
0.5252 0.6347 0.6880 0.6490 0.7100 0.7478 0.7958 0.7595 0.8149 0.0981 0.0514 0.0843 0.0658 0.0922 0.1139 0.0529 0.1113 0.0846 0.1321 0.0897
0.4967 0.5864 0.6611 0.6384 0.6439 0.7252 0.7579 0.7433 0.7643 0.7761 0.0757 0.0799 0.0757 0.0864 0.0959 0.0685 0.1234 0.0733 0.1842 0.0930
0.4475 0.5803 0.6970 0.6617 0.6656 0.7274 0.7326 0.7567 0.7699 0.7763 0.8150 0.0869 0.0600 0.1025 0.1103 0.0537 0.1281 0.1026 0.2042 0.0885
0.5281 0.5656 0.6930 0.5233 0.6089 0.6503 0.7072 0.7421 0.7283 0.7588 0.7663 0.7865 0.0570 0.1119 0.1056 0.0495 0.1348 0.0943 0.1699 0.0789
0.5007 0.5260 0.6689 0.5819 0.5772 0.5277 0.6928 0.7165 0.7096 0.7229 0.7449 0.7382 0.8271 0.1264 0.1036 0.0706 0.1131 0.0985 0.1906 0.0820
0.4961 0.4526 0.6794 0.5465 0.5234 0.5238 0.6348 0.6833 0.6908 0.6886 0.7163 0.7292 0.7810 0.7952 0.1053 0.0661 0.1135 0.0942 0.2164 0.0806
0.4743 0.3400 0.5585 0.5611 0.5606 0.5228 0.6122 0.5523 0.6552 0.6740 0.6804 0.7231 0.7709 0.7340 0.8112 0.0747 0.1137 0.0811 0.1933 0.0988
0.4730 0.3533 0.5391 0.4235 0.4580 0.4516 0.6392 0.4629 0.6467 0.6725 0.6265 0.7099 0.7861 0.7021 0.7690 0.8066 0.0756 0.0953 0.1826 0.1147
0.4648 0.3295 0.5033 0.3883 0.4151 0.4542 0.5422 0.4495 0.5977 0.6719 0.5681 0.6966 0.7251 0.6391 0.7507 0.7260 0.7925 0.1094 0.1796 0.1389
0.4468 0.3238 0.4941 0.3709 0.4352 0.4632 0.5464 0.4534 0.6094 0.6366 0.6202 0.6787 0.6911 0.6615 0.7107 0.7206 0.8035 0.8109 0.1575 0.1526
0.4513 0.2867 0.4935 0.3826 0.4880 0.4347 0.5075 0.4067 0.5570 0.5492 0.5810 0.6007 0.6724 0.6000 0.7521 0.6801 0.7754 0.7353 0.8174 0.1532
0.4690 0.3482 0.5277 0.3742 0.4973 0.4406 0.5743 0.4568 0.6527 0.5518 0.6356 0.6396 0.6216 0.6537 0.7388 0.7072 0.8005 0.7300 0.8062 0.8107

Final Accuracy: 0.6018
Backward: -0.1980
Forward:  0.0093
\end{Verbatim}}
\subsubsection{Model \emph{independent}}
{\tiny
\begin{Verbatim}
0.0936 0.0995 0.0884 0.0893 0.0784 0.1000 0.1108 0.0965 0.1243 0.1048 0.0819 0.1115 0.0999 0.0762 0.1060 0.1260 0.0930 0.1075 0.1126 0.1092
|
0.1821 0.0995 0.0884 0.0893 0.0784 0.1000 0.1108 0.0965 0.1243 0.1048 0.0819 0.1115 0.0999 0.0762 0.1060 0.1260 0.0930 0.1075 0.1126 0.1092
0.1821 0.2601 0.0884 0.0893 0.0784 0.1000 0.1108 0.0965 0.1243 0.1048 0.0819 0.1115 0.0999 0.0762 0.1060 0.1260 0.0930 0.1075 0.1126 0.1092
0.1821 0.2601 0.3538 0.0893 0.0784 0.1000 0.1108 0.0965 0.1243 0.1048 0.0819 0.1115 0.0999 0.0762 0.1060 0.1260 0.0930 0.1075 0.1126 0.1092
0.1821 0.2601 0.3538 0.3089 0.0784 0.1000 0.1108 0.0965 0.1243 0.1048 0.0819 0.1115 0.0999 0.0762 0.1060 0.1260 0.0930 0.1075 0.1126 0.1092
0.1821 0.2601 0.3538 0.3089 0.4031 0.1000 0.1108 0.0965 0.1243 0.1048 0.0819 0.1115 0.0999 0.0762 0.1060 0.1260 0.0930 0.1075 0.1126 0.1092
0.1821 0.2601 0.3538 0.3089 0.4031 0.3267 0.1108 0.0965 0.1243 0.1048 0.0819 0.1115 0.0999 0.0762 0.1060 0.1260 0.0930 0.1075 0.1126 0.1092
0.1821 0.2601 0.3538 0.3089 0.4031 0.3267 0.4071 0.0965 0.1243 0.1048 0.0819 0.1115 0.0999 0.0762 0.1060 0.1260 0.0930 0.1075 0.1126 0.1092
0.1821 0.2601 0.3538 0.3089 0.4031 0.3267 0.4071 0.4030 0.1243 0.1048 0.0819 0.1115 0.0999 0.0762 0.1060 0.1260 0.0930 0.1075 0.1126 0.1092
0.1821 0.2601 0.3538 0.3089 0.4031 0.3267 0.4071 0.4030 0.4495 0.1048 0.0819 0.1115 0.0999 0.0762 0.1060 0.1260 0.0930 0.1075 0.1126 0.1092
0.1821 0.2601 0.3538 0.3089 0.4031 0.3267 0.4071 0.4030 0.4495 0.5276 0.0819 0.1115 0.0999 0.0762 0.1060 0.1260 0.0930 0.1075 0.1126 0.1092
0.1821 0.2601 0.3538 0.3089 0.4031 0.3267 0.4071 0.4030 0.4495 0.5276 0.4779 0.1115 0.0999 0.0762 0.1060 0.1260 0.0930 0.1075 0.1126 0.1092
0.1821 0.2601 0.3538 0.3089 0.4031 0.3267 0.4071 0.4030 0.4495 0.5276 0.4779 0.5357 0.0999 0.0762 0.1060 0.1260 0.0930 0.1075 0.1126 0.1092
0.1821 0.2601 0.3538 0.3089 0.4031 0.3267 0.4071 0.4030 0.4495 0.5276 0.4779 0.5357 0.5393 0.0762 0.1060 0.1260 0.0930 0.1075 0.1126 0.1092
0.1821 0.2601 0.3538 0.3089 0.4031 0.3267 0.4071 0.4030 0.4495 0.5276 0.4779 0.5357 0.5393 0.5755 0.1060 0.1260 0.0930 0.1075 0.1126 0.1092
0.1821 0.2601 0.3538 0.3089 0.4031 0.3267 0.4071 0.4030 0.4495 0.5276 0.4779 0.5357 0.5393 0.5755 0.5592 0.1260 0.0930 0.1075 0.1126 0.1092
0.1821 0.2601 0.3538 0.3089 0.4031 0.3267 0.4071 0.4030 0.4495 0.5276 0.4779 0.5357 0.5393 0.5755 0.5592 0.5145 0.0930 0.1075 0.1126 0.1092
0.1821 0.2601 0.3538 0.3089 0.4031 0.3267 0.4071 0.4030 0.4495 0.5276 0.4779 0.5357 0.5393 0.5755 0.5592 0.5145 0.5553 0.1075 0.1126 0.1092
0.1821 0.2601 0.3538 0.3089 0.4031 0.3267 0.4071 0.4030 0.4495 0.5276 0.4779 0.5357 0.5393 0.5755 0.5592 0.5145 0.5553 0.5675 0.1126 0.1092
0.1821 0.2601 0.3538 0.3089 0.4031 0.3267 0.4071 0.4030 0.4495 0.5276 0.4779 0.5357 0.5393 0.5755 0.5592 0.5145 0.5553 0.5675 0.5252 0.1092
0.1821 0.2601 0.3538 0.3089 0.4031 0.3267 0.4071 0.4030 0.4495 0.5276 0.4779 0.5357 0.5393 0.5755 0.5592 0.5145 0.5553 0.5675 0.5252 0.5739

Final Accuracy: 0.4523
Backward: 0.0000
Forward:  0.0000
\end{Verbatim}}
\newpage
\subsubsection{Model \emph{multimodal}}
{\tiny
\begin{Verbatim}
0.0749 0.1152 0.0601 0.0885 0.0826 0.0856 0.0925 0.0703 0.1079 0.0891 0.1029 0.1092 0.0866 0.1014 0.1575 0.1005 0.1083 0.1038 0.0857 0.0759
|
0.6871 0.1306 0.1402 0.1209 0.1108 0.1135 0.1195 0.1066 0.1368 0.1217 0.0605 0.1168 0.1094 0.0970 0.1150 0.1322 0.1471 0.1243 0.1054 0.1567
0.7123 0.8211 0.1344 0.1197 0.0923 0.1155 0.1402 0.1030 0.1346 0.1424 0.1030 0.1329 0.0910 0.0875 0.1437 0.1245 0.1094 0.0865 0.0771 0.1768
0.7716 0.8180 0.8163 0.1208 0.0871 0.1154 0.1308 0.0916 0.1482 0.1238 0.1092 0.1159 0.1122 0.0868 0.1192 0.1221 0.1518 0.0981 0.0692 0.1362
0.7451 0.7961 0.7976 0.7757 0.0883 0.1152 0.1235 0.0849 0.1612 0.1265 0.0976 0.1177 0.1112 0.0914 0.1177 0.1072 0.1681 0.0847 0.0724 0.1495
0.6841 0.7759 0.7569 0.7709 0.7887 0.1147 0.1296 0.0749 0.1561 0.1259 0.0958 0.1216 0.1104 0.0878 0.1261 0.1410 0.1496 0.1009 0.0741 0.1783
0.7344 0.8274 0.8126 0.7832 0.8041 0.8057 0.1145 0.0647 0.1449 0.1446 0.0826 0.1047 0.0862 0.0896 0.1324 0.1362 0.1217 0.0773 0.0878 0.1640
0.7679 0.8373 0.8228 0.7891 0.8117 0.7830 0.8464 0.0673 0.1467 0.1437 0.0831 0.1015 0.0864 0.0880 0.1340 0.1264 0.1212 0.0763 0.0819 0.1532
0.7327 0.8004 0.7749 0.7608 0.7858 0.8231 0.8253 0.7830 0.1472 0.1343 0.1082 0.1209 0.1179 0.1115 0.1159 0.1263 0.1357 0.0862 0.0988 0.1824
0.7606 0.7713 0.7570 0.7962 0.7871 0.8118 0.8393 0.7462 0.7877 0.1288 0.0877 0.1018 0.0819 0.0664 0.1287 0.1060 0.1107 0.0679 0.0801 0.1169
0.7213 0.7791 0.7889 0.7818 0.7401 0.7756 0.8084 0.7793 0.7996 0.7521 0.0575 0.1151 0.0893 0.0667 0.1280 0.1092 0.1194 0.0717 0.0732 0.1365
0.6891 0.7691 0.7889 0.7772 0.7392 0.8103 0.7917 0.7584 0.8111 0.7634 0.8079 0.0984 0.0758 0.0815 0.1398 0.1195 0.1019 0.0849 0.0783 0.1104
0.7357 0.8039 0.7822 0.7710 0.7673 0.8090 0.7918 0.8125 0.7698 0.7689 0.7859 0.7383 0.0858 0.0681 0.1369 0.1135 0.1170 0.0762 0.0780 0.1330
0.7082 0.7856 0.7612 0.7351 0.7639 0.7871 0.7741 0.8013 0.7273 0.7818 0.7458 0.6813 0.8387 0.0607 0.1368 0.1084 0.1106 0.0686 0.0696 0.1195
0.7344 0.7662 0.7261 0.7582 0.7844 0.8025 0.8005 0.7910 0.7397 0.7952 0.7445 0.6660 0.8303 0.6274 0.1447 0.1076 0.1078 0.0711 0.0655 0.1039
0.7347 0.7716 0.7370 0.7814 0.7901 0.7909 0.8188 0.7887 0.7521 0.7739 0.7428 0.6771 0.8373 0.6218 0.8413 0.1006 0.0890 0.0667 0.0688 0.0922
0.7544 0.7908 0.7497 0.8055 0.7880 0.8026 0.8243 0.7854 0.7665 0.7741 0.7767 0.7199 0.8211 0.6111 0.8349 0.7976 0.0985 0.0813 0.0811 0.0934
0.7579 0.7631 0.7103 0.7976 0.7695 0.7876 0.8066 0.7507 0.7860 0.7708 0.7680 0.6620 0.7952 0.5357 0.8083 0.7646 0.7769 0.0813 0.0730 0.0990
0.6829 0.7751 0.7282 0.7604 0.7526 0.7427 0.7484 0.7788 0.7369 0.7357 0.7413 0.6109 0.8131 0.5895 0.8224 0.7750 0.8211 0.7871 0.0771 0.0881
0.7128 0.7917 0.7226 0.7779 0.7647 0.7361 0.8046 0.7454 0.7549 0.7559 0.7757 0.6850 0.8092 0.5882 0.8022 0.7528 0.8176 0.7988 0.7844 0.0902
0.7068 0.7832 0.7106 0.7704 0.7642 0.7382 0.7866 0.7218 0.7623 0.7496 0.7829 0.7072 0.8015 0.5605 0.7974 0.7645 0.8115 0.8045 0.7904 0.8077

Final Accuracy: 0.7561
Backward: -0.0275
Forward:  0.0059
\end{Verbatim}}
\subsubsection{Model \emph{EWC}}
{\tiny
\begin{Verbatim}
0.1330 0.1199 0.1070 0.0825 0.0609 0.0832 0.1385 0.1123 0.0736 0.1190 0.0666 0.0890 0.0885 0.0723 0.1083 0.0524 0.0976 0.0871 0.1143 0.0743
|
0.7289 0.1426 0.0776 0.1272 0.0637 0.0582 0.0944 0.0961 0.0691 0.0794 0.0822 0.1426 0.0953 0.0792 0.0885 0.1094 0.1328 0.0987 0.1507 0.0976
0.7396 0.8305 0.0703 0.1585 0.0749 0.0545 0.1558 0.0789 0.1076 0.0857 0.0699 0.1361 0.0953 0.0581 0.0704 0.0850 0.1241 0.1280 0.1382 0.1012
0.6882 0.7731 0.8274 0.1326 0.0496 0.0917 0.1571 0.0767 0.1015 0.1061 0.0790 0.0712 0.0622 0.0632 0.0982 0.0803 0.1535 0.1088 0.1399 0.1134
0.6361 0.7054 0.7915 0.7780 0.0733 0.0681 0.1446 0.0587 0.0888 0.1046 0.1141 0.0731 0.1008 0.0928 0.0968 0.1183 0.1263 0.1112 0.1130 0.1141
0.6625 0.7168 0.7724 0.7663 0.7665 0.0807 0.1015 0.0576 0.1105 0.0735 0.0783 0.0703 0.0951 0.0744 0.0658 0.0775 0.1378 0.0938 0.1127 0.1226
0.6667 0.6854 0.7791 0.7800 0.7795 0.7990 0.0850 0.0597 0.1037 0.0543 0.0515 0.0714 0.0566 0.1021 0.0906 0.0791 0.1603 0.0997 0.1140 0.1108
0.6193 0.6901 0.7954 0.7871 0.7436 0.7666 0.8404 0.0960 0.1232 0.0752 0.0660 0.0898 0.0460 0.0707 0.1058 0.0696 0.1410 0.1144 0.0948 0.1175
0.5454 0.6454 0.7321 0.7166 0.7331 0.7609 0.8361 0.7918 0.1004 0.0805 0.0540 0.0902 0.0451 0.1068 0.0924 0.0720 0.1259 0.1018 0.0988 0.1001
0.5594 0.6857 0.7414 0.6629 0.7248 0.6523 0.7804 0.7862 0.7973 0.0906 0.0790 0.0921 0.0481 0.1199 0.0959 0.0650 0.1147 0.0790 0.1291 0.1143
0.5880 0.6145 0.7121 0.7267 0.6943 0.6630 0.7527 0.7069 0.7303 0.7930 0.0978 0.0607 0.0488 0.1357 0.0969 0.0989 0.1437 0.0632 0.1132 0.1148
0.5455 0.6000 0.7029 0.6885 0.6554 0.6926 0.6618 0.7296 0.7446 0.7713 0.8005 0.0739 0.0583 0.1253 0.1082 0.0656 0.1396 0.1021 0.1393 0.1066
0.5770 0.6049 0.6553 0.5227 0.6169 0.5683 0.6525 0.5805 0.6772 0.6869 0.7180 0.7147 0.0421 0.1338 0.0964 0.0602 0.1148 0.1037 0.1060 0.0878
0.5668 0.6023 0.6847 0.6257 0.5574 0.6179 0.6576 0.5748 0.5917 0.7406 0.7422 0.7544 0.8085 0.1488 0.1065 0.0981 0.0996 0.0986 0.0949 0.0722
0.5325 0.5731 0.6141 0.4784 0.5395 0.5387 0.6384 0.5310 0.6705 0.6309 0.6515 0.7252 0.7926 0.7110 0.1178 0.0927 0.0911 0.1137 0.0887 0.0977
0.5039 0.5394 0.6175 0.5436 0.5573 0.5783 0.5886 0.4870 0.6058 0.6844 0.6531 0.7011 0.7850 0.6647 0.8235 0.0619 0.0990 0.0922 0.0998 0.1172
0.5799 0.4831 0.5988 0.4822 0.5807 0.4963 0.6100 0.4051 0.6081 0.6858 0.5913 0.7020 0.7989 0.6506 0.7815 0.8030 0.0902 0.0877 0.0912 0.1250
0.5399 0.4113 0.5397 0.4447 0.5240 0.5427 0.5627 0.3753 0.5512 0.6142 0.4868 0.7001 0.7361 0.6206 0.6931 0.7176 0.7411 0.1161 0.1373 0.1347
0.5427 0.5133 0.5789 0.5249 0.6101 0.5422 0.5875 0.4113 0.6144 0.5876 0.5636 0.6325 0.7244 0.6313 0.6796 0.7146 0.7886 0.7659 0.1224 0.1180
0.5085 0.5428 0.6279 0.5754 0.5676 0.5175 0.6138 0.4533 0.6017 0.5632 0.6322 0.5923 0.6966 0.6234 0.6764 0.6761 0.7688 0.7045 0.7836 0.1122
0.4839 0.5742 0.5863 0.5711 0.5760 0.5578 0.5912 0.4302 0.6911 0.5543 0.5981 0.6211 0.6812 0.6415 0.6357 0.5967 0.7709 0.7132 0.7234 0.7715

Final Accuracy: 0.6185
Backward: -0.1653
Forward:  0.0054
\end{Verbatim}}
\subsubsection{Model \emph{GEM}}
{\tiny
\begin{Verbatim}
0.1330 0.1199 0.1070 0.0825 0.0609 0.0832 0.1385 0.1123 0.0736 0.1190 0.0666 0.0890 0.0885 0.0723 0.1083 0.0524 0.0976 0.0871 0.1143 0.0743
|
0.7289 0.1426 0.0776 0.1272 0.0637 0.0582 0.0944 0.0961 0.0691 0.0794 0.0822 0.1426 0.0953 0.0792 0.0885 0.1094 0.1328 0.0987 0.1507 0.0976
0.8158 0.8356 0.0601 0.1668 0.0708 0.0624 0.1524 0.0836 0.0994 0.0757 0.0802 0.1278 0.1056 0.0634 0.0625 0.0924 0.1193 0.1390 0.1392 0.1031
0.8134 0.8302 0.8407 0.1509 0.0539 0.0987 0.1566 0.0785 0.1134 0.1025 0.0659 0.0808 0.0746 0.0593 0.0883 0.0759 0.1764 0.1191 0.1253 0.1067
0.7732 0.7940 0.8306 0.7605 0.0622 0.0789 0.1374 0.0669 0.1172 0.0735 0.0979 0.0488 0.0877 0.0802 0.0713 0.1161 0.1434 0.1216 0.1295 0.1032
0.8079 0.8283 0.8478 0.8328 0.7798 0.0797 0.0918 0.0598 0.1376 0.0753 0.0704 0.0516 0.0986 0.0573 0.0549 0.0802 0.1815 0.1077 0.1249 0.1279
0.8262 0.8296 0.8296 0.8471 0.8472 0.8093 0.0888 0.0661 0.0942 0.0890 0.0572 0.0487 0.0707 0.0954 0.0957 0.0793 0.1793 0.1086 0.1306 0.1049
0.8109 0.8214 0.8132 0.8374 0.8445 0.8429 0.8297 0.0984 0.0999 0.0971 0.0699 0.0608 0.0662 0.0722 0.0910 0.0814 0.1720 0.1175 0.1218 0.1165
0.8098 0.8185 0.8268 0.8388 0.8275 0.8391 0.8587 0.7566 0.1019 0.1059 0.0751 0.0727 0.0563 0.1015 0.1116 0.0856 0.1505 0.0991 0.1793 0.1144
0.8260 0.8247 0.8447 0.8290 0.8468 0.8362 0.8517 0.8246 0.8219 0.1096 0.0734 0.0662 0.0595 0.1137 0.0923 0.0855 0.1417 0.1073 0.1398 0.1156
0.8202 0.8264 0.8237 0.8271 0.8462 0.8458 0.8462 0.8307 0.8574 0.8142 0.0670 0.0787 0.0691 0.0957 0.0950 0.0975 0.1403 0.0828 0.1587 0.1204
0.8213 0.8137 0.8228 0.8257 0.8352 0.8517 0.8422 0.8296 0.8515 0.8559 0.8041 0.0808 0.0628 0.1036 0.1005 0.0805 0.1351 0.1107 0.1509 0.1026
0.7878 0.8073 0.8088 0.7913 0.8250 0.8290 0.8328 0.8196 0.8388 0.8343 0.8172 0.7556 0.0686 0.1228 0.0976 0.0922 0.1308 0.1180 0.1360 0.0870
0.8114 0.8181 0.8149 0.8128 0.8328 0.8273 0.8407 0.8254 0.8487 0.8551 0.8339 0.8550 0.8224 0.1212 0.1038 0.1207 0.1185 0.0969 0.1461 0.0928
0.7982 0.8191 0.8133 0.8094 0.8350 0.8343 0.8495 0.8186 0.8473 0.8566 0.8279 0.8552 0.8505 0.7850 0.1130 0.1232 0.1264 0.0816 0.1328 0.1051
0.8044 0.8217 0.8170 0.8198 0.8429 0.8240 0.8490 0.8147 0.8498 0.8568 0.8334 0.8551 0.8489 0.8442 0.8291 0.1096 0.1197 0.0890 0.1746 0.1045
0.8150 0.8198 0.8126 0.8250 0.8348 0.8281 0.8476 0.8191 0.8487 0.8551 0.8343 0.8575 0.8520 0.8447 0.8517 0.8155 0.1148 0.0896 0.1269 0.1269
0.8100 0.8164 0.8030 0.8162 0.8355 0.8242 0.8455 0.8149 0.8348 0.8531 0.8311 0.8541 0.8528 0.8411 0.8558 0.8367 0.8253 0.1083 0.1452 0.1035
0.8031 0.8092 0.8047 0.8170 0.8294 0.8103 0.8421 0.8087 0.8384 0.8479 0.8379 0.8399 0.8467 0.8413 0.8478 0.8377 0.8501 0.8287 0.1335 0.1102
0.7999 0.8107 0.8038 0.8033 0.8168 0.8067 0.8433 0.8044 0.8309 0.8397 0.8355 0.8349 0.8419 0.8320 0.8379 0.8363 0.8421 0.8369 0.7689 0.1119
0.8024 0.8063 0.7966 0.8045 0.8199 0.8153 0.8430 0.8009 0.8317 0.8494 0.8304 0.8391 0.8431 0.8341 0.8426 0.8226 0.8461 0.8436 0.8350 0.8141

Final Accuracy: 0.8260
Backward: 0.0247
Forward:  0.0088
\end{Verbatim}}

\subsection{MNIST rotations}
\subsubsection{Model \emph{single}}
{\tiny
\begin{Verbatim}
0.0903 0.0957 0.0843 0.0882 0.0910 0.0847 0.0998 0.0992 0.0906 0.0756 0.0743 0.0781 0.0839 0.0873 0.0778 0.0810 0.0780 0.0791 0.0884 0.0867
|
0.2777 0.2559 0.2387 0.2382 0.2057 0.1597 0.1548 0.1284 0.1106 0.0912 0.0881 0.0889 0.1082 0.1030 0.0950 0.1011 0.1059 0.1130 0.1263 0.1490
0.4166 0.4466 0.3915 0.3657 0.3048 0.2447 0.2062 0.1701 0.1476 0.1358 0.1285 0.1236 0.1473 0.1221 0.1164 0.1212 0.1074 0.1174 0.1394 0.1646
0.4598 0.5531 0.5332 0.4883 0.4369 0.3510 0.2676 0.2098 0.1822 0.1580 0.1467 0.1379 0.1568 0.1279 0.1240 0.1280 0.1167 0.1215 0.1426 0.1642
0.4760 0.5879 0.5823 0.5779 0.5265 0.4513 0.3448 0.2649 0.2321 0.1786 0.1593 0.1441 0.1541 0.1301 0.1241 0.1240 0.1234 0.1334 0.1529 0.1677
0.4619 0.6080 0.6358 0.6383 0.6069 0.5365 0.4179 0.3122 0.2670 0.1996 0.1727 0.1534 0.1486 0.1271 0.1290 0.1296 0.1319 0.1426 0.1701 0.1892
0.4522 0.6096 0.6652 0.6865 0.6793 0.6383 0.5041 0.3951 0.3331 0.2389 0.1984 0.1681 0.1609 0.1388 0.1443 0.1396 0.1465 0.1489 0.1731 0.1916
0.4191 0.5759 0.6415 0.6779 0.6902 0.6793 0.6057 0.5175 0.4500 0.3221 0.2554 0.1986 0.1720 0.1410 0.1420 0.1406 0.1486 0.1547 0.1770 0.2025
0.4160 0.5657 0.6248 0.6618 0.7004 0.7245 0.6977 0.6576 0.6200 0.4765 0.3857 0.2642 0.2134 0.1595 0.1524 0.1504 0.1471 0.1562 0.1726 0.1965
0.3671 0.5139 0.5866 0.6303 0.6699 0.7021 0.7016 0.6990 0.6878 0.5734 0.4855 0.3443 0.2578 0.1729 0.1503 0.1441 0.1446 0.1525 0.1738 0.1936
0.3269 0.4819 0.5596 0.6152 0.6666 0.7057 0.7219 0.7343 0.7272 0.6526 0.5712 0.4225 0.3191 0.2081 0.1758 0.1648 0.1479 0.1543 0.1665 0.1796
0.3098 0.4567 0.5334 0.5945 0.6441 0.6876 0.7189 0.7482 0.7576 0.7061 0.6630 0.5231 0.4143 0.2643 0.2052 0.1907 0.1557 0.1607 0.1692 0.1877
0.2879 0.4181 0.4965 0.5614 0.6175 0.6678 0.7104 0.7449 0.7595 0.7329 0.7159 0.6327 0.5347 0.3718 0.2821 0.2556 0.1917 0.1882 0.1873 0.1977
0.2797 0.4064 0.4695 0.5275 0.5915 0.6495 0.6925 0.7310 0.7486 0.7341 0.7398 0.6979 0.6550 0.5034 0.3892 0.3615 0.2490 0.2378 0.2121 0.2079
0.2545 0.3551 0.4139 0.4677 0.5308 0.5871 0.6519 0.6939 0.7186 0.7197 0.7367 0.7369 0.7174 0.6270 0.5141 0.4821 0.3397 0.3041 0.2487 0.2346
0.2613 0.3557 0.3999 0.4496 0.5069 0.5534 0.6194 0.6540 0.6815 0.6885 0.7134 0.7317 0.7440 0.7064 0.6553 0.6275 0.4707 0.4288 0.3217 0.2804
0.2485 0.3330 0.3686 0.4169 0.4739 0.5138 0.5891 0.6245 0.6503 0.6715 0.6947 0.7258 0.7536 0.7440 0.7201 0.7051 0.5773 0.5326 0.3963 0.3444
0.2468 0.3107 0.3275 0.3603 0.4160 0.4402 0.5168 0.5672 0.5917 0.6307 0.6577 0.7025 0.7393 0.7559 0.7551 0.7474 0.6757 0.6328 0.5115 0.4579
0.2243 0.2914 0.3098 0.3437 0.3873 0.4008 0.4748 0.5190 0.5470 0.5950 0.6142 0.6730 0.7173 0.7530 0.7723 0.7738 0.7373 0.7128 0.5946 0.5345
0.2055 0.2771 0.2953 0.3319 0.3715 0.3848 0.4420 0.4860 0.5122 0.5549 0.5759 0.6355 0.6768 0.7149 0.7563 0.7592 0.7544 0.7478 0.6940 0.6390
0.2136 0.2710 0.2837 0.3126 0.3460 0.3508 0.4033 0.4461 0.4702 0.5237 0.5408 0.6021 0.6513 0.6987 0.7481 0.7590 0.7708 0.7772 0.7434 0.7023

Final Accuracy: 0.5307
Backward: -0.0896
Forward:  0.4254
\end{Verbatim}}
\subsubsection{Model \emph{independent}}
{\tiny
\begin{Verbatim}
0.1013 0.0910 0.1049 0.0761 0.0976 0.1071 0.0960 0.0890 0.0815 0.1172 0.0938 0.1231 0.1140 0.0874 0.0995 0.1005 0.1075 0.0919 0.0942 0.1790
|
0.4221 0.0910 0.1049 0.0761 0.0976 0.1071 0.0960 0.0890 0.0815 0.1172 0.0938 0.1231 0.1140 0.0874 0.0995 0.1005 0.1075 0.0919 0.0942 0.1790
0.4221 0.5266 0.1049 0.0761 0.0976 0.1071 0.0960 0.0890 0.0815 0.1172 0.0938 0.1231 0.1140 0.0874 0.0995 0.1005 0.1075 0.0919 0.0942 0.1790
0.4221 0.5266 0.5069 0.0761 0.0976 0.1071 0.0960 0.0890 0.0815 0.1172 0.0938 0.1231 0.1140 0.0874 0.0995 0.1005 0.1075 0.0919 0.0942 0.1790
0.4221 0.5266 0.5069 0.5967 0.0976 0.1071 0.0960 0.0890 0.0815 0.1172 0.0938 0.1231 0.1140 0.0874 0.0995 0.1005 0.1075 0.0919 0.0942 0.1790
0.4221 0.5266 0.5069 0.5967 0.5091 0.1071 0.0960 0.0890 0.0815 0.1172 0.0938 0.1231 0.1140 0.0874 0.0995 0.1005 0.1075 0.0919 0.0942 0.1790
0.4221 0.5266 0.5069 0.5967 0.5091 0.6481 0.0960 0.0890 0.0815 0.1172 0.0938 0.1231 0.1140 0.0874 0.0995 0.1005 0.1075 0.0919 0.0942 0.1790
0.4221 0.5266 0.5069 0.5967 0.5091 0.6481 0.6278 0.0890 0.0815 0.1172 0.0938 0.1231 0.1140 0.0874 0.0995 0.1005 0.1075 0.0919 0.0942 0.1790
0.4221 0.5266 0.5069 0.5967 0.5091 0.6481 0.6278 0.5654 0.0815 0.1172 0.0938 0.1231 0.1140 0.0874 0.0995 0.1005 0.1075 0.0919 0.0942 0.1790
0.4221 0.5266 0.5069 0.5967 0.5091 0.6481 0.6278 0.5654 0.5874 0.1172 0.0938 0.1231 0.1140 0.0874 0.0995 0.1005 0.1075 0.0919 0.0942 0.1790
0.4221 0.5266 0.5069 0.5967 0.5091 0.6481 0.6278 0.5654 0.5874 0.5711 0.0938 0.1231 0.1140 0.0874 0.0995 0.1005 0.1075 0.0919 0.0942 0.1790
0.4221 0.5266 0.5069 0.5967 0.5091 0.6481 0.6278 0.5654 0.5874 0.5711 0.6846 0.1231 0.1140 0.0874 0.0995 0.1005 0.1075 0.0919 0.0942 0.1790
0.4221 0.5266 0.5069 0.5967 0.5091 0.6481 0.6278 0.5654 0.5874 0.5711 0.6846 0.6723 0.1140 0.0874 0.0995 0.1005 0.1075 0.0919 0.0942 0.1790
0.4221 0.5266 0.5069 0.5967 0.5091 0.6481 0.6278 0.5654 0.5874 0.5711 0.6846 0.6723 0.6533 0.0874 0.0995 0.1005 0.1075 0.0919 0.0942 0.1790
0.4221 0.5266 0.5069 0.5967 0.5091 0.6481 0.6278 0.5654 0.5874 0.5711 0.6846 0.6723 0.6533 0.5998 0.0995 0.1005 0.1075 0.0919 0.0942 0.1790
0.4221 0.5266 0.5069 0.5967 0.5091 0.6481 0.6278 0.5654 0.5874 0.5711 0.6846 0.6723 0.6533 0.5998 0.7046 0.1005 0.1075 0.0919 0.0942 0.1790
0.4221 0.5266 0.5069 0.5967 0.5091 0.6481 0.6278 0.5654 0.5874 0.5711 0.6846 0.6723 0.6533 0.5998 0.7046 0.6965 0.1075 0.0919 0.0942 0.1790
0.4221 0.5266 0.5069 0.5967 0.5091 0.6481 0.6278 0.5654 0.5874 0.5711 0.6846 0.6723 0.6533 0.5998 0.7046 0.6965 0.7011 0.0919 0.0942 0.1790
0.4221 0.5266 0.5069 0.5967 0.5091 0.6481 0.6278 0.5654 0.5874 0.5711 0.6846 0.6723 0.6533 0.5998 0.7046 0.6965 0.7011 0.7532 0.0942 0.1790
0.4221 0.5266 0.5069 0.5967 0.5091 0.6481 0.6278 0.5654 0.5874 0.5711 0.6846 0.6723 0.6533 0.5998 0.7046 0.6965 0.7011 0.7532 0.7424 0.1790
0.4221 0.5266 0.5069 0.5967 0.5091 0.6481 0.6278 0.5654 0.5874 0.5711 0.6846 0.6723 0.6533 0.5998 0.7046 0.6965 0.7011 0.7532 0.7424 0.7131

Final Accuracy: 0.6241
Backward: 0.0000
Forward:  0.0000
\end{Verbatim}}
\subsubsection{Model \emph{multimodal}}
{\tiny
\begin{Verbatim}
0.0910 0.0870 0.0943 0.0850 0.1043 0.1102 0.0820 0.0696 0.1339 0.0871 0.0946 0.0901 0.0961 0.0506 0.0991 0.0819 0.1018 0.0846 0.0778 0.1116
|
0.6383 0.1237 0.1273 0.1238 0.1322 0.1184 0.1173 0.1289 0.1152 0.1068 0.1253 0.1499 0.1202 0.1165 0.1327 0.1088 0.1264 0.0729 0.1397 0.1097
0.6893 0.8122 0.1502 0.1124 0.1077 0.1306 0.1124 0.1104 0.1749 0.0445 0.1239 0.1555 0.0726 0.1460 0.1044 0.0768 0.1344 0.0667 0.0936 0.1140
0.7519 0.7990 0.7989 0.1042 0.0798 0.1415 0.0978 0.0843 0.1692 0.0656 0.1042 0.1048 0.0994 0.1430 0.0601 0.0966 0.1270 0.0688 0.0754 0.1310
0.7601 0.7709 0.7155 0.7898 0.0860 0.1465 0.0952 0.0756 0.1799 0.0509 0.1106 0.1279 0.1067 0.1275 0.0607 0.0947 0.1293 0.0746 0.0888 0.1442
0.6484 0.7890 0.7966 0.8229 0.7745 0.1345 0.0966 0.1006 0.1790 0.0381 0.1081 0.1532 0.1041 0.1488 0.0745 0.0627 0.1352 0.0588 0.1351 0.1240
0.7081 0.7838 0.8047 0.8242 0.7983 0.8155 0.1031 0.1075 0.1842 0.0345 0.0954 0.1556 0.1202 0.1501 0.0727 0.0764 0.1280 0.0617 0.1047 0.1041
0.7024 0.7591 0.8092 0.8247 0.7894 0.8185 0.8534 0.0859 0.1825 0.0374 0.0848 0.1266 0.1013 0.1776 0.0483 0.0851 0.1186 0.0583 0.1074 0.0956
0.6513 0.7391 0.7529 0.8224 0.7751 0.7736 0.7943 0.8205 0.1899 0.0575 0.1190 0.1675 0.1221 0.1437 0.1153 0.0963 0.1484 0.0676 0.1175 0.1182
0.6982 0.7214 0.7352 0.8050 0.7633 0.7760 0.7679 0.7914 0.7938 0.0640 0.0981 0.1094 0.1005 0.1551 0.0467 0.1016 0.1342 0.0753 0.0884 0.1025
0.7137 0.7398 0.7571 0.8232 0.7566 0.8061 0.8019 0.8078 0.7705 0.7824 0.1036 0.1169 0.1033 0.1426 0.0397 0.0948 0.1914 0.0688 0.0853 0.1082
0.6852 0.7492 0.7546 0.8247 0.7722 0.8114 0.8113 0.7814 0.7905 0.7918 0.8244 0.1497 0.1274 0.1331 0.0725 0.0907 0.1988 0.0564 0.0846 0.0966
0.6730 0.7220 0.7138 0.7417 0.7212 0.8037 0.7353 0.7522 0.7307 0.8016 0.7666 0.7150 0.1374 0.1520 0.0688 0.0977 0.1953 0.0512 0.0916 0.0855
0.6708 0.7275 0.7316 0.7717 0.7189 0.8170 0.7439 0.7635 0.7686 0.7958 0.7602 0.7145 0.8476 0.1562 0.0531 0.0954 0.1447 0.0565 0.0712 0.0676
0.7254 0.6934 0.6972 0.7631 0.7215 0.7612 0.7491 0.7587 0.7416 0.7863 0.7467 0.6608 0.8258 0.6332 0.0521 0.0971 0.1256 0.0805 0.0856 0.0970
0.7337 0.6945 0.7183 0.7578 0.7289 0.7751 0.7479 0.7891 0.7405 0.8019 0.7546 0.6818 0.8461 0.6459 0.8378 0.1011 0.1223 0.0799 0.0763 0.0945
0.7242 0.6911 0.7217 0.7509 0.7299 0.7866 0.7370 0.7865 0.7550 0.8301 0.7717 0.6701 0.8528 0.6085 0.8414 0.8089 0.1177 0.0786 0.0751 0.0997
0.7510 0.7034 0.7479 0.7737 0.7465 0.7734 0.7507 0.8094 0.7728 0.8020 0.7903 0.6261 0.8367 0.6149 0.8214 0.7959 0.7532 0.0788 0.0762 0.0928
0.7208 0.6948 0.7312 0.7529 0.7133 0.7838 0.7536 0.8028 0.7382 0.7821 0.7730 0.6797 0.8350 0.6902 0.8278 0.7852 0.7852 0.7722 0.0750 0.0758
0.6976 0.7028 0.7227 0.7413 0.6956 0.7953 0.7429 0.8143 0.7494 0.7935 0.7792 0.6846 0.8354 0.6694 0.8254 0.7987 0.7890 0.7792 0.7901 0.0834
0.7065 0.7008 0.7246 0.7463 0.6862 0.7965 0.7323 0.8055 0.7398 0.7872 0.7736 0.6983 0.8370 0.6515 0.8280 0.7981 0.7994 0.7817 0.7832 0.7928

Final Accuracy: 0.7585
Backward: -0.0243
Forward:  0.0177
\end{Verbatim}}
\subsubsection{Model \emph{EWC}}
{\tiny
\begin{Verbatim}
0.0903 0.0957 0.0843 0.0882 0.0910 0.0847 0.0998 0.0992 0.0906 0.0756 0.0743 0.0781 0.0839 0.0873 0.0778 0.0810 0.0780 0.0791 0.0884 0.0867
|
0.5388 0.4506 0.3594 0.3211 0.2657 0.2081 0.1768 0.1425 0.1281 0.1229 0.1250 0.1321 0.1629 0.1452 0.1390 0.1385 0.1452 0.1508 0.1690 0.1831
0.6677 0.7185 0.6303 0.5681 0.4726 0.3893 0.2711 0.2194 0.2009 0.1661 0.1614 0.1541 0.1547 0.1378 0.1429 0.1309 0.1450 0.1541 0.1775 0.1850
0.6407 0.7390 0.7261 0.6820 0.6034 0.5138 0.3681 0.2737 0.2352 0.1812 0.1661 0.1429 0.1408 0.1319 0.1411 0.1309 0.1466 0.1582 0.1779 0.1820
0.6090 0.7484 0.7841 0.7763 0.7212 0.6385 0.4619 0.3382 0.2771 0.2069 0.1822 0.1554 0.1438 0.1369 0.1459 0.1389 0.1507 0.1570 0.1688 0.1775
0.5821 0.7637 0.8103 0.8164 0.7878 0.7165 0.5352 0.3950 0.3269 0.2367 0.2023 0.1794 0.1677 0.1517 0.1552 0.1510 0.1532 0.1621 0.1781 0.1934
0.4946 0.6821 0.7669 0.8024 0.8114 0.7989 0.6709 0.5325 0.4530 0.3242 0.2628 0.2027 0.1822 0.1570 0.1621 0.1524 0.1534 0.1571 0.1695 0.1820
0.4532 0.6274 0.7089 0.7662 0.7960 0.8216 0.7669 0.6809 0.6191 0.4623 0.3647 0.2488 0.2023 0.1575 0.1548 0.1517 0.1474 0.1533 0.1665 0.1791
0.4296 0.6019 0.6951 0.7478 0.7900 0.8227 0.8156 0.7810 0.7416 0.6254 0.5120 0.3423 0.2714 0.1869 0.1704 0.1602 0.1467 0.1483 0.1484 0.1661
0.3748 0.5299 0.6121 0.6724 0.7289 0.7833 0.8095 0.8095 0.7960 0.7063 0.6105 0.4368 0.3335 0.2148 0.1842 0.1712 0.1496 0.1505 0.1548 0.1645
0.3254 0.4824 0.5717 0.6362 0.7059 0.7562 0.7980 0.8112 0.8081 0.7510 0.6698 0.5132 0.3878 0.2416 0.1922 0.1819 0.1400 0.1382 0.1339 0.1250
0.3239 0.4702 0.5456 0.6064 0.6671 0.7282 0.7805 0.8154 0.8253 0.8043 0.7706 0.6440 0.5137 0.3283 0.2432 0.2252 0.1649 0.1624 0.1561 0.1542
0.2904 0.4132 0.4724 0.5268 0.5831 0.6426 0.7224 0.7650 0.7902 0.7957 0.8104 0.7633 0.6818 0.5040 0.3822 0.3468 0.2227 0.2033 0.1803 0.1648
0.2627 0.3784 0.4305 0.4844 0.5428 0.5957 0.6715 0.7177 0.7479 0.7716 0.8070 0.8100 0.7724 0.6127 0.4755 0.4429 0.2755 0.2451 0.1860 0.1691
0.2650 0.3579 0.3955 0.4304 0.4782 0.5086 0.5763 0.6263 0.6576 0.6866 0.7372 0.7834 0.8041 0.7512 0.6558 0.6225 0.4341 0.3840 0.2818 0.2343
0.2537 0.3630 0.3971 0.4268 0.4656 0.4821 0.5349 0.5703 0.6037 0.6249 0.6776 0.7357 0.7648 0.7700 0.7470 0.7243 0.5666 0.5086 0.3507 0.2739
0.2460 0.3470 0.3780 0.4074 0.4503 0.4661 0.5114 0.5514 0.5837 0.6105 0.6547 0.7185 0.7585 0.7834 0.7835 0.7782 0.6416 0.5873 0.4241 0.3461
0.2838 0.3438 0.3453 0.3670 0.4010 0.4097 0.4292 0.4794 0.5080 0.5416 0.5934 0.6701 0.7147 0.7673 0.7947 0.7951 0.7438 0.7013 0.5586 0.4891
0.2606 0.3373 0.3489 0.3761 0.4059 0.4112 0.4184 0.4572 0.4866 0.5167 0.5591 0.6303 0.6715 0.7450 0.7909 0.7981 0.7877 0.7728 0.6445 0.5752
0.2598 0.3731 0.4139 0.4396 0.4562 0.4614 0.4630 0.4731 0.4872 0.4995 0.5168 0.5576 0.5771 0.6386 0.6787 0.6911 0.7179 0.7388 0.7337 0.6951
0.2645 0.3793 0.4190 0.4424 0.4693 0.4561 0.4453 0.4455 0.4568 0.4852 0.4992 0.5384 0.5713 0.6441 0.6865 0.6998 0.7426 0.7718 0.7628 0.7414

Final Accuracy: 0.5461
Backward: -0.2047
Forward:  0.5524
\end{Verbatim}}
\subsubsection{Model \emph{GEM}}
{\tiny
\begin{Verbatim}
0.0903 0.0957 0.0843 0.0882 0.0910 0.0847 0.0998 0.0992 0.0906 0.0756 0.0743 0.0781 0.0839 0.0873 0.0778 0.0810 0.0780 0.0791 0.0884 0.0867
|
0.7150 0.6438 0.5319 0.4525 0.3707 0.2810 0.1841 0.1309 0.1217 0.1064 0.1065 0.0997 0.0981 0.1055 0.1258 0.1306 0.1641 0.1670 0.1745 0.1746
0.8393 0.8542 0.7718 0.6760 0.5440 0.4304 0.2996 0.2332 0.2152 0.1786 0.1528 0.1297 0.1214 0.1141 0.1268 0.1263 0.1457 0.1592 0.1762 0.1835
0.7994 0.8744 0.8766 0.8386 0.7664 0.6496 0.4421 0.3067 0.2596 0.1915 0.1770 0.1524 0.1490 0.1369 0.1377 0.1357 0.1352 0.1463 0.1414 0.1411
0.8051 0.8883 0.9008 0.8941 0.8503 0.7569 0.5371 0.3685 0.2996 0.2025 0.1667 0.1246 0.1204 0.1062 0.1078 0.1074 0.1174 0.1330 0.1446 0.1531
0.7618 0.8691 0.8882 0.8906 0.8648 0.7938 0.6290 0.4748 0.4036 0.2955 0.2262 0.1661 0.1494 0.1258 0.1334 0.1309 0.1464 0.1583 0.1852 0.1990
0.7982 0.8797 0.9024 0.9039 0.8983 0.8786 0.7858 0.6460 0.5636 0.3800 0.2922 0.1947 0.1705 0.1360 0.1335 0.1306 0.1327 0.1476 0.1699 0.1763
0.7691 0.8492 0.8718 0.8825 0.8834 0.8860 0.8516 0.7770 0.7165 0.5728 0.4280 0.2638 0.2115 0.1425 0.1273 0.1196 0.1041 0.1144 0.1240 0.1389
0.7970 0.8704 0.8938 0.9050 0.9076 0.9135 0.9090 0.8788 0.8498 0.7472 0.6185 0.3839 0.2985 0.1820 0.1537 0.1479 0.1410 0.1439 0.1556 0.1626
0.7733 0.8433 0.8623 0.8763 0.8785 0.8909 0.8900 0.8895 0.8762 0.8190 0.7209 0.5179 0.4132 0.2511 0.1965 0.1865 0.1594 0.1626 0.1621 0.1705
0.8019 0.8729 0.8920 0.9083 0.9062 0.9059 0.9074 0.9033 0.9012 0.8710 0.7910 0.6302 0.4971 0.3003 0.2310 0.2054 0.1636 0.1657 0.1760 0.1859
0.7525 0.8296 0.8507 0.8591 0.8494 0.8492 0.8479 0.8473 0.8460 0.8233 0.8029 0.6590 0.5487 0.3576 0.2717 0.2369 0.1646 0.1594 0.1586 0.1659
0.7470 0.8471 0.8726 0.8832 0.8893 0.8901 0.8932 0.9009 0.9029 0.9076 0.9006 0.8677 0.8089 0.6113 0.4586 0.4161 0.2529 0.2181 0.1831 0.1994
0.7846 0.8612 0.8801 0.8879 0.8874 0.8910 0.8886 0.8959 0.8972 0.8971 0.9064 0.8994 0.8789 0.7509 0.6201 0.5783 0.3766 0.3191 0.2496 0.2356
0.7565 0.8396 0.8660 0.8766 0.8768 0.8783 0.8732 0.8657 0.8646 0.8535 0.8506 0.8571 0.8651 0.8403 0.7835 0.7558 0.5672 0.4996 0.3637 0.3021
0.6846 0.7966 0.8435 0.8663 0.8794 0.8858 0.8868 0.8891 0.8900 0.8844 0.8854 0.8901 0.8952 0.8884 0.8486 0.8184 0.6492 0.5637 0.3904 0.3136
0.7074 0.8058 0.8376 0.8509 0.8513 0.8507 0.8555 0.8575 0.8599 0.8513 0.8539 0.8665 0.8735 0.8757 0.8726 0.8651 0.7586 0.6984 0.5102 0.4074
0.7536 0.8433 0.8657 0.8772 0.8779 0.8779 0.8804 0.8891 0.8928 0.8850 0.8872 0.8869 0.8929 0.8980 0.8933 0.8921 0.8508 0.8196 0.6974 0.5975
0.7565 0.8486 0.8726 0.8781 0.8760 0.8669 0.8700 0.8702 0.8687 0.8578 0.8578 0.8644 0.8730 0.8902 0.8983 0.8995 0.8872 0.8728 0.7845 0.6955
0.7259 0.8321 0.8560 0.8643 0.8649 0.8498 0.8539 0.8634 0.8595 0.8484 0.8412 0.8532 0.8579 0.8719 0.8892 0.8956 0.8968 0.8966 0.8622 0.8130
0.7295 0.8222 0.8484 0.8594 0.8603 0.8568 0.8586 0.8645 0.8654 0.8626 0.8650 0.8621 0.8703 0.8697 0.8817 0.8843 0.8948 0.9046 0.8860 0.8669

Final Accuracy: 0.8607
Backward: 0.0048
Forward:  0.6647
\end{Verbatim}}

\subsection{CIFAR-100 incremental}
\subsubsection{Model \emph{single}}
{\tiny
\begin{Verbatim}
0.2000 0.2000 0.2000 0.2000 0.1980 0.2000 0.1980 0.1980 0.2000 0.2000 0.2000 0.2000 0.2000 0.2000 0.2000 0.2000 0.2000 0.1980 0.2000 0.2000
|
0.3080 0.1760 0.1580 0.2240 0.1940 0.1960 0.2620 0.2160 0.2020 0.2280 0.1920 0.2280 0.1960 0.1880 0.1680 0.1320 0.2100 0.1940 0.1580 0.2620
0.2860 0.2340 0.1900 0.2260 0.1540 0.2040 0.1620 0.1880 0.2000 0.1960 0.2800 0.1880 0.2360 0.1440 0.1320 0.1260 0.1640 0.1280 0.0880 0.2020
0.3800 0.2440 0.4000 0.2040 0.1840 0.2060 0.1920 0.1100 0.1980 0.2420 0.1900 0.2060 0.2560 0.1560 0.0980 0.1180 0.2040 0.1840 0.1240 0.2600
0.2600 0.1980 0.3480 0.4620 0.2000 0.1920 0.1560 0.1540 0.1960 0.2240 0.2000 0.2420 0.3720 0.1420 0.1480 0.1180 0.1360 0.1900 0.1800 0.1920
0.2060 0.2540 0.3780 0.4220 0.6520 0.2120 0.1240 0.1600 0.2020 0.1780 0.2320 0.1940 0.2400 0.1680 0.1520 0.0820 0.1620 0.1580 0.2160 0.2140
0.2880 0.2500 0.3200 0.3500 0.6480 0.4200 0.1020 0.2020 0.1860 0.1800 0.1480 0.1980 0.2780 0.1180 0.1060 0.1500 0.1860 0.1700 0.2040 0.2200
0.2980 0.2600 0.3260 0.3260 0.4620 0.3260 0.6220 0.1820 0.2100 0.1460 0.1880 0.2740 0.2560 0.1100 0.1140 0.2200 0.2020 0.1880 0.1920 0.0960
0.2680 0.2040 0.3580 0.3820 0.4140 0.3760 0.6020 0.4920 0.2020 0.2120 0.1820 0.2020 0.2860 0.1260 0.0920 0.1480 0.2180 0.1540 0.2100 0.2300
0.3120 0.1980 0.3140 0.3780 0.4300 0.3920 0.5320 0.4500 0.5580 0.1780 0.2920 0.1860 0.2680 0.1280 0.1460 0.2200 0.1880 0.1480 0.1480 0.1940
0.2360 0.2020 0.3940 0.3300 0.4100 0.3860 0.5000 0.3660 0.4000 0.6640 0.2260 0.2220 0.2380 0.1540 0.1080 0.1340 0.1780 0.1220 0.2160 0.3520
0.2740 0.2180 0.2360 0.3440 0.4260 0.3180 0.4040 0.4340 0.3580 0.5860 0.8060 0.1920 0.2580 0.1360 0.1860 0.1260 0.1960 0.1460 0.2520 0.2660
0.3460 0.2440 0.3220 0.3280 0.3840 0.3100 0.4800 0.4260 0.4300 0.5980 0.6780 0.5760 0.2740 0.2000 0.1480 0.1380 0.2300 0.1400 0.1460 0.3140
0.2780 0.2240 0.3500 0.4200 0.4320 0.3540 0.5040 0.4040 0.3060 0.5420 0.6240 0.5060 0.6740 0.1720 0.1560 0.1100 0.2340 0.1980 0.2900 0.1960
0.1940 0.2640 0.3000 0.3380 0.4520 0.3040 0.3920 0.3940 0.3840 0.5840 0.6140 0.4160 0.6020 0.6440 0.1300 0.1380 0.2120 0.1940 0.3220 0.1980
0.2180 0.2840 0.2140 0.3180 0.4820 0.2780 0.4420 0.4160 0.3340 0.4860 0.6000 0.4820 0.6300 0.6240 0.7560 0.1900 0.2340 0.1880 0.3040 0.2360
0.2580 0.2720 0.2780 0.3040 0.4620 0.3060 0.4300 0.3880 0.4020 0.4760 0.4760 0.5320 0.5940 0.5040 0.6840 0.6240 0.2380 0.1680 0.3200 0.3620
0.2440 0.2600 0.2620 0.3840 0.3760 0.2600 0.3760 0.3280 0.2660 0.3900 0.5120 0.5260 0.5540 0.4500 0.6520 0.5200 0.6760 0.1700 0.3280 0.2720
0.2060 0.2520 0.3140 0.3660 0.4380 0.2880 0.3760 0.3820 0.3880 0.4300 0.5400 0.5200 0.6420 0.4660 0.6620 0.5420 0.6020 0.6600 0.2520 0.2080
0.2300 0.2780 0.3200 0.3800 0.4520 0.2760 0.3580 0.4080 0.4040 0.4800 0.5700 0.5300 0.6340 0.4860 0.7240 0.5860 0.5700 0.5640 0.7540 0.2900
0.2560 0.2520 0.3020 0.3900 0.4000 0.2740 0.3220 0.3760 0.3300 0.5240 0.5860 0.4680 0.5980 0.4380 0.6980 0.5140 0.6180 0.4840 0.7000 0.7320

Final Accuracy: 0.4631
Backward: -0.1226
Forward:  -0.0006
\end{Verbatim}}
\subsubsection{Model \emph{independent}}
{\tiny
\begin{Verbatim}
0.2000 0.2000 0.2000 0.1980 0.2000 0.1980 0.1980 0.2000 0.1980 0.2000 0.2000 0.2000 0.2000 0.2000 0.1980 0.2000 0.1980 0.2000 0.2000 0.2000
|
0.4180 0.2000 0.2000 0.1980 0.2000 0.1980 0.1980 0.2000 0.1980 0.2000 0.2000 0.2000 0.2000 0.2000 0.1980 0.2000 0.1980 0.2000 0.2000 0.2000
0.4180 0.4140 0.2000 0.1980 0.2000 0.1980 0.1980 0.2000 0.1980 0.2000 0.2000 0.2000 0.2000 0.2000 0.1980 0.2000 0.1980 0.2000 0.2000 0.2000
0.4180 0.4140 0.3640 0.1980 0.2000 0.1980 0.1980 0.2000 0.1980 0.2000 0.2000 0.2000 0.2000 0.2000 0.1980 0.2000 0.1980 0.2000 0.2000 0.2000
0.4180 0.4140 0.3640 0.3200 0.2000 0.1980 0.1980 0.2000 0.1980 0.2000 0.2000 0.2000 0.2000 0.2000 0.1980 0.2000 0.1980 0.2000 0.2000 0.2000
0.4180 0.4140 0.3640 0.3200 0.5620 0.1980 0.1980 0.2000 0.1980 0.2000 0.2000 0.2000 0.2000 0.2000 0.1980 0.2000 0.1980 0.2000 0.2000 0.2000
0.4180 0.4140 0.3640 0.3200 0.5620 0.3000 0.1980 0.2000 0.1980 0.2000 0.2000 0.2000 0.2000 0.2000 0.1980 0.2000 0.1980 0.2000 0.2000 0.2000
0.4180 0.4140 0.3640 0.3200 0.5620 0.3000 0.4100 0.2000 0.1980 0.2000 0.2000 0.2000 0.2000 0.2000 0.1980 0.2000 0.1980 0.2000 0.2000 0.2000
0.4180 0.4140 0.3640 0.3200 0.5620 0.3000 0.4100 0.2880 0.1980 0.2000 0.2000 0.2000 0.2000 0.2000 0.1980 0.2000 0.1980 0.2000 0.2000 0.2000
0.4180 0.4140 0.3640 0.3200 0.5620 0.3000 0.4100 0.2880 0.3440 0.2000 0.2000 0.2000 0.2000 0.2000 0.1980 0.2000 0.1980 0.2000 0.2000 0.2000
0.4180 0.4140 0.3640 0.3200 0.5620 0.3000 0.4100 0.2880 0.3440 0.3880 0.2000 0.2000 0.2000 0.2000 0.1980 0.2000 0.1980 0.2000 0.2000 0.2000
0.4180 0.4140 0.3640 0.3200 0.5620 0.3000 0.4100 0.2880 0.3440 0.3880 0.6120 0.2000 0.2000 0.2000 0.1980 0.2000 0.1980 0.2000 0.2000 0.2000
0.4180 0.4140 0.3640 0.3200 0.5620 0.3000 0.4100 0.2880 0.3440 0.3880 0.6120 0.4800 0.2000 0.2000 0.1980 0.2000 0.1980 0.2000 0.2000 0.2000
0.4180 0.4140 0.3640 0.3200 0.5620 0.3000 0.4100 0.2880 0.3440 0.3880 0.6120 0.4800 0.5060 0.2000 0.1980 0.2000 0.1980 0.2000 0.2000 0.2000
0.4180 0.4140 0.3640 0.3200 0.5620 0.3000 0.4100 0.2880 0.3440 0.3880 0.6120 0.4800 0.5060 0.5180 0.1980 0.2000 0.1980 0.2000 0.2000 0.2000
0.4180 0.4140 0.3640 0.3200 0.5620 0.3000 0.4100 0.2880 0.3440 0.3880 0.6120 0.4800 0.5060 0.5180 0.4740 0.2000 0.1980 0.2000 0.2000 0.2000
0.4180 0.4140 0.3640 0.3200 0.5620 0.3000 0.4100 0.2880 0.3440 0.3880 0.6120 0.4800 0.5060 0.5180 0.4740 0.4520 0.1980 0.2000 0.2000 0.2000
0.4180 0.4140 0.3640 0.3200 0.5620 0.3000 0.4100 0.2880 0.3440 0.3880 0.6120 0.4800 0.5060 0.5180 0.4740 0.4520 0.4160 0.2000 0.2000 0.2000
0.4180 0.4140 0.3640 0.3200 0.5620 0.3000 0.4100 0.2880 0.3440 0.3880 0.6120 0.4800 0.5060 0.5180 0.4740 0.4520 0.4160 0.3620 0.2000 0.2000
0.4180 0.4140 0.3640 0.3200 0.5620 0.3000 0.4100 0.2880 0.3440 0.3880 0.6120 0.4800 0.5060 0.5180 0.4740 0.4520 0.4160 0.3620 0.4340 0.2000
0.4180 0.4140 0.3640 0.3200 0.5620 0.3000 0.4100 0.2880 0.3440 0.3880 0.6120 0.4800 0.5060 0.5180 0.4740 0.4520 0.4160 0.3620 0.4340 0.4080

Final Accuracy: 0.4235
Backward: 0.0000
Forward:  0.0000
\end{Verbatim}}
\subsubsection{Model \emph{iCARL}}
{\tiny
\begin{Verbatim}
0.2000 0.2000 0.2000 0.2000 0.2000 0.2000 0.1980 0.2000 0.2000 0.2000 0.2000 0.2000 0.2000 0.1980 0.1980 0.1980 0.2000 0.2000 0.1980 0.2000
|
0.3560 0.2000 0.2000 0.2000 0.2000 0.2000 0.1980 0.2000 0.2000 0.2000 0.2000 0.2000 0.2000 0.1980 0.1980 0.1980 0.2000 0.2000 0.1980 0.2000
0.3880 0.5060 0.2000 0.2000 0.2000 0.2000 0.1980 0.2000 0.2000 0.2000 0.2000 0.2000 0.2000 0.1980 0.1980 0.1980 0.2000 0.2000 0.1980 0.2000
0.3840 0.4060 0.5040 0.2000 0.2000 0.2000 0.1980 0.2000 0.2000 0.2000 0.2000 0.2000 0.2000 0.1980 0.1980 0.1980 0.2000 0.2000 0.1980 0.2000
0.3500 0.3900 0.5100 0.5740 0.2000 0.2000 0.1980 0.2000 0.2000 0.2000 0.2000 0.2000 0.2000 0.1980 0.1980 0.1980 0.2000 0.2000 0.1980 0.2000
0.4160 0.3760 0.4320 0.4660 0.6320 0.2000 0.1980 0.2000 0.2000 0.2000 0.2000 0.2000 0.2000 0.1980 0.1980 0.1980 0.2000 0.2000 0.1980 0.2000
0.4820 0.4760 0.4340 0.4940 0.6040 0.5320 0.1980 0.2000 0.2000 0.2000 0.2000 0.2000 0.2000 0.1980 0.1980 0.1980 0.2000 0.2000 0.1980 0.2000
0.4620 0.4240 0.4080 0.4900 0.5540 0.3840 0.6380 0.2000 0.2000 0.2000 0.2000 0.2000 0.2000 0.1980 0.1980 0.1980 0.2000 0.2000 0.1980 0.2000
0.4880 0.4360 0.4600 0.4940 0.5480 0.4520 0.5240 0.5600 0.2000 0.2000 0.2000 0.2000 0.2000 0.1980 0.1980 0.1980 0.2000 0.2000 0.1980 0.2000
0.4620 0.4560 0.3820 0.4480 0.5200 0.4000 0.4800 0.4840 0.5520 0.2000 0.2000 0.2000 0.2000 0.1980 0.1980 0.1980 0.2000 0.2000 0.1980 0.2000
0.4900 0.4480 0.4600 0.4400 0.5340 0.3780 0.4760 0.4620 0.4500 0.7540 0.2000 0.2000 0.2000 0.1980 0.1980 0.1980 0.2000 0.2000 0.1980 0.2000
0.4740 0.4520 0.3700 0.4480 0.4960 0.3680 0.4400 0.5200 0.4480 0.6120 0.7620 0.2000 0.2000 0.1980 0.1980 0.1980 0.2000 0.2000 0.1980 0.2000
0.4400 0.4640 0.3940 0.4480 0.4840 0.3900 0.4680 0.4660 0.4320 0.5620 0.6240 0.5940 0.2000 0.1980 0.1980 0.1980 0.2000 0.2000 0.1980 0.2000
0.5080 0.4740 0.4460 0.4880 0.5020 0.4220 0.5080 0.4940 0.4700 0.5560 0.6680 0.5200 0.7340 0.1980 0.1980 0.1980 0.2000 0.2000 0.1980 0.2000
0.4640 0.4540 0.4320 0.4760 0.5660 0.4240 0.4920 0.4860 0.3860 0.5340 0.6560 0.4480 0.6200 0.6060 0.1980 0.1980 0.2000 0.2000 0.1980 0.2000
0.5360 0.4580 0.3940 0.4380 0.5720 0.4120 0.5140 0.4720 0.4040 0.5040 0.5980 0.4660 0.6140 0.5580 0.7300 0.1980 0.2000 0.2000 0.1980 0.2000
0.5040 0.5280 0.4680 0.4640 0.5460 0.4280 0.5300 0.5020 0.4800 0.5960 0.6660 0.5340 0.6020 0.5120 0.6680 0.6960 0.2000 0.2000 0.1980 0.2000
0.5160 0.5080 0.4800 0.4700 0.5620 0.4040 0.4180 0.5060 0.4700 0.5860 0.6860 0.5320 0.6580 0.4660 0.6940 0.5960 0.7140 0.2000 0.1980 0.2000
0.5580 0.4580 0.4400 0.5320 0.5220 0.4460 0.5000 0.5020 0.4880 0.5480 0.6680 0.5000 0.6560 0.4640 0.5660 0.5000 0.6200 0.6900 0.1980 0.2000
0.5600 0.5160 0.4480 0.4940 0.5780 0.4300 0.4520 0.4900 0.4820 0.5440 0.7160 0.4960 0.6620 0.4760 0.6620 0.4920 0.5760 0.5680 0.7200 0.2000
0.5160 0.4820 0.4240 0.5020 0.6060 0.4260 0.5320 0.5380 0.4620 0.6060 0.6680 0.4920 0.6420 0.4460 0.6600 0.5060 0.5600 0.5280 0.5980 0.7300

Final Accuracy: 0.5462
Backward: -0.0830
Forward:  0.0000
\end{Verbatim}}
\subsubsection{Model \emph{EWC}}
{\tiny
\begin{Verbatim}
0.2000 0.2000 0.2000 0.2000 0.1980 0.2000 0.1980 0.1980 0.2000 0.2000 0.2000 0.2000 0.2000 0.2000 0.2000 0.2000 0.2000 0.1980 0.2000 0.2000
|
0.3080 0.1760 0.1580 0.2240 0.1940 0.1960 0.2620 0.2160 0.2020 0.2280 0.1920 0.2280 0.1960 0.1880 0.1680 0.1320 0.2100 0.1940 0.1580 0.2620
0.2960 0.3380 0.2100 0.2320 0.1380 0.2380 0.1580 0.2260 0.1800 0.2080 0.2080 0.2100 0.2320 0.1820 0.2000 0.1760 0.1800 0.1700 0.1380 0.2220
0.3480 0.2700 0.4340 0.2840 0.2120 0.2080 0.2200 0.0800 0.2000 0.2480 0.2560 0.2020 0.3260 0.2040 0.2380 0.1020 0.2040 0.1820 0.1880 0.1220
0.3040 0.2660 0.4140 0.4840 0.1820 0.2200 0.2180 0.1020 0.1380 0.2360 0.0680 0.2100 0.2440 0.2140 0.1840 0.1220 0.1300 0.1260 0.2880 0.0740
0.2660 0.2060 0.3800 0.4160 0.6040 0.2180 0.1220 0.1400 0.2020 0.2080 0.2120 0.2220 0.1860 0.2900 0.2880 0.1500 0.1780 0.1780 0.1920 0.1160
0.2400 0.2320 0.3880 0.4340 0.4440 0.3460 0.1080 0.1080 0.1560 0.1940 0.2020 0.2020 0.1560 0.1860 0.2680 0.1340 0.1940 0.1700 0.2520 0.1960
0.2720 0.2560 0.3160 0.3820 0.4580 0.2800 0.6100 0.1440 0.1960 0.2040 0.0940 0.2000 0.1900 0.2080 0.3280 0.1540 0.2060 0.1800 0.2640 0.2240
0.3080 0.2840 0.4340 0.5140 0.5340 0.3800 0.5480 0.4720 0.2300 0.2300 0.0800 0.2120 0.2100 0.2740 0.3760 0.0720 0.1920 0.1880 0.2480 0.2560
0.3480 0.2820 0.3600 0.4480 0.4420 0.4240 0.4040 0.3720 0.5440 0.1920 0.1160 0.2460 0.1440 0.2800 0.2120 0.0680 0.1500 0.1860 0.2500 0.2220
0.2680 0.2300 0.3500 0.4400 0.3780 0.4100 0.4340 0.3840 0.4200 0.6560 0.2400 0.2020 0.1480 0.2220 0.2720 0.1700 0.1980 0.1820 0.2060 0.3140
0.3060 0.2760 0.3400 0.4300 0.3900 0.4120 0.3860 0.3920 0.3880 0.5700 0.7860 0.1520 0.2040 0.2380 0.1980 0.1400 0.2160 0.1900 0.2120 0.2000
0.3420 0.2920 0.2980 0.3540 0.3380 0.3780 0.4860 0.3860 0.4500 0.5940 0.6600 0.6240 0.1120 0.2100 0.2080 0.1540 0.2100 0.1780 0.1380 0.1820
0.3080 0.2300 0.3700 0.3940 0.4120 0.4020 0.3380 0.3620 0.4120 0.5980 0.5400 0.4920 0.6720 0.2600 0.1880 0.1120 0.2040 0.1840 0.1620 0.1880
0.2240 0.2280 0.3340 0.4540 0.4640 0.4040 0.4240 0.3820 0.3720 0.5520 0.5620 0.4540 0.6140 0.5840 0.1660 0.1160 0.1840 0.2000 0.1860 0.2080
0.2540 0.2140 0.3240 0.3920 0.5200 0.3520 0.3880 0.4580 0.3620 0.5400 0.5500 0.4500 0.5700 0.5380 0.7300 0.1340 0.1640 0.1960 0.1540 0.2300
0.3040 0.2540 0.2780 0.3340 0.4240 0.2840 0.3720 0.3320 0.4240 0.4580 0.4740 0.4320 0.4480 0.4480 0.5280 0.5480 0.1540 0.2060 0.2220 0.2880
0.3300 0.2240 0.3360 0.4480 0.4440 0.4280 0.3280 0.3620 0.3620 0.4540 0.6120 0.5300 0.5940 0.4440 0.6120 0.5500 0.6940 0.1980 0.1720 0.2140
0.3460 0.2400 0.3120 0.4440 0.3460 0.3460 0.3740 0.3960 0.4140 0.5240 0.5900 0.5040 0.6440 0.4720 0.5780 0.5260 0.5400 0.6720 0.2620 0.2160
0.2980 0.2140 0.4500 0.4640 0.3960 0.4420 0.3680 0.4340 0.4080 0.5260 0.5740 0.4760 0.6540 0.5340 0.6980 0.5220 0.6360 0.5080 0.7560 0.2160
0.4220 0.2160 0.3580 0.4240 0.3120 0.3980 0.4440 0.4260 0.4120 0.5880 0.6500 0.4640 0.6480 0.5180 0.6980 0.5300 0.5320 0.5180 0.7060 0.7040

Final Accuracy: 0.4984
Backward: -0.0799
Forward:  -0.0077
\end{Verbatim}}
\subsubsection{Model \emph{GEM}}
{\tiny
\begin{Verbatim}
0.2000 0.2000 0.2000 0.2000 0.1980 0.2000 0.1980 0.1980 0.2000 0.2000 0.2000 0.2000 0.2000 0.2000 0.2000 0.2000 0.2000 0.1980 0.2000 0.2000
|
0.5680 0.1720 0.2080 0.2620 0.2900 0.2000 0.2100 0.2160 0.1800 0.2000 0.2120 0.2200 0.2480 0.2240 0.2280 0.2600 0.2320 0.1840 0.1840 0.1600
0.5340 0.4540 0.2200 0.2600 0.2700 0.2100 0.2380 0.1680 0.2020 0.1940 0.2440 0.2080 0.2540 0.1940 0.2920 0.1860 0.1560 0.1960 0.1560 0.1540
0.6280 0.4900 0.4860 0.2300 0.2760 0.2040 0.1740 0.1520 0.2620 0.2300 0.2300 0.1900 0.2880 0.1220 0.2460 0.1680 0.1640 0.1760 0.2040 0.1460
0.6320 0.5960 0.5640 0.6020 0.2880 0.2140 0.2300 0.1340 0.2160 0.2200 0.2040 0.2180 0.1960 0.1320 0.2420 0.2340 0.1580 0.1560 0.1960 0.2120
0.5160 0.5720 0.5540 0.6420 0.7520 0.2040 0.1820 0.1500 0.2520 0.1640 0.1580 0.1620 0.2200 0.1460 0.2100 0.1480 0.1360 0.1480 0.1840 0.1800
0.5020 0.5780 0.5600 0.5620 0.8040 0.6140 0.1580 0.2280 0.2860 0.1920 0.2120 0.2180 0.2300 0.1340 0.2000 0.1280 0.0660 0.1380 0.2160 0.1840
0.5500 0.5400 0.5460 0.5700 0.7620 0.5140 0.7340 0.1520 0.2400 0.1560 0.1760 0.1720 0.2660 0.1500 0.2140 0.1680 0.1200 0.1380 0.2560 0.1800
0.5300 0.6140 0.6080 0.6020 0.7840 0.5800 0.6680 0.5780 0.2620 0.1940 0.1640 0.1740 0.2960 0.1260 0.2020 0.1680 0.0720 0.1820 0.2160 0.1600
0.5000 0.6020 0.6040 0.6420 0.7780 0.5940 0.6400 0.6480 0.6400 0.1760 0.1460 0.1860 0.3020 0.1200 0.1780 0.1460 0.1600 0.1420 0.1840 0.1620
0.5640 0.6020 0.5860 0.6320 0.7100 0.5720 0.6900 0.6280 0.6320 0.7740 0.1840 0.1900 0.2640 0.1640 0.1380 0.1720 0.1360 0.1400 0.1960 0.1400
0.5800 0.6260 0.5940 0.6280 0.7360 0.5460 0.6200 0.6100 0.6240 0.7340 0.7720 0.1740 0.3000 0.1480 0.2060 0.1420 0.1020 0.1620 0.1520 0.1520
0.5480 0.6340 0.6200 0.6040 0.7460 0.5340 0.6840 0.6160 0.6460 0.7520 0.7640 0.6880 0.2540 0.1600 0.1860 0.1480 0.0960 0.1380 0.1800 0.1580
0.5620 0.6020 0.6480 0.6280 0.7300 0.5940 0.6760 0.6500 0.6340 0.7460 0.7520 0.7100 0.7480 0.1720 0.1800 0.1640 0.0900 0.1700 0.1620 0.1700
0.5440 0.5760 0.6020 0.6060 0.7120 0.5480 0.6720 0.6380 0.6160 0.7240 0.7220 0.6560 0.7780 0.6940 0.1820 0.1740 0.1060 0.1440 0.1960 0.2200
0.5660 0.6140 0.6040 0.6280 0.7340 0.6180 0.6940 0.6100 0.6380 0.7080 0.7200 0.6580 0.7700 0.7140 0.7740 0.1900 0.1120 0.1540 0.1640 0.2060
0.5120 0.5760 0.5940 0.6360 0.7280 0.5660 0.7220 0.6340 0.5840 0.6960 0.6920 0.6120 0.7720 0.6920 0.7580 0.6340 0.1140 0.1240 0.1840 0.1840
0.5840 0.6140 0.6520 0.6340 0.7580 0.5820 0.6740 0.6660 0.6360 0.7340 0.7160 0.6720 0.7460 0.7000 0.7360 0.6820 0.7300 0.1360 0.1880 0.1780
0.5800 0.6100 0.6440 0.6500 0.6800 0.5820 0.7020 0.6440 0.5780 0.7200 0.7340 0.6700 0.7200 0.6760 0.7300 0.7040 0.7620 0.6980 0.1780 0.1860
0.6020 0.6180 0.6640 0.6500 0.7440 0.5980 0.6660 0.6100 0.5900 0.7160 0.7360 0.6580 0.7700 0.6900 0.7400 0.6960 0.7240 0.6720 0.8200 0.1900
0.5640 0.6300 0.6620 0.6780 0.7160 0.5720 0.6740 0.5920 0.6140 0.7320 0.7200 0.6720 0.7640 0.6760 0.7320 0.6860 0.7180 0.6340 0.8080 0.7220

Final Accuracy: 0.6783
Backward: 0.0042
Forward:  -0.0078
\end{Verbatim}}

% \end{document}

%% file: continuum.bbl
\begin{thebibliography}{53}
\providecommand{\natexlab}[1]{#1}
\providecommand{\url}[1]{\texttt{#1}}
\expandafter\ifx\csname urlstyle\endcsname\relax
  \providecommand{\doi}[1]{doi: #1}\else
  \providecommand{\doi}{doi: \begingroup \urlstyle{rm}\Url}\fi

\bibitem[{Aljundi} et~al.(2016){Aljundi}, {Chakravarty}, and
  {Tuytelaars}]{expert_gate}
R.~{Aljundi}, P.~{Chakravarty}, and T.~{Tuytelaars}.
\newblock Expert gate: Lifelong learning with a network of experts.
\newblock \emph{CVPR}, 2016.

\bibitem[Balcan et~al.(2015)Balcan, Blum, and Vempola]{balcan_lifelong}
M.-F. Balcan, A.~Blum, and S.~Vempola.
\newblock Efficient representations for lifelong learning and autoencoding.
\newblock \emph{COLT}, 2015.

\bibitem[{Baroni} et~al.(2017){Baroni}, {Joulin}, {Jabri}, {Kruszewski},
  {Lazaridou}, {Simonic}, and {Mikolov}]{commai2}
M.~{Baroni}, A.~{Joulin}, A.~{Jabri}, G.~{Kruszewski}, A.~{Lazaridou},
  K.~{Simonic}, and T.~{Mikolov}.
\newblock {CommAI: Evaluating the first steps towards a useful general AI}.
\newblock \emph{arXiv}, 2017.

\bibitem[Baxter(2000)]{baxter}
J.~Baxter.
\newblock A model of inductive bias learning.
\newblock \emph{JAIR}, 2000.

\bibitem[Ben-David et~al.(2010)Ben-David, Blitzer, Crammer, Kulesza, Pereira,
  and Wortman~Vaughan]{domain_adaptation}
S.~Ben-David, J.~Blitzer, K.~Crammer, A.~Kulesza, F.~Pereira, and
  J.~Wortman~Vaughan.
\newblock A theory of learning from different domains.
\newblock \emph{Machine Learning Journal}, 2010.

\bibitem[Bengio et~al.(2009)Bengio, Louradour, Collobert, and
  Weston]{curriculum}
Y.~Bengio, J.~Louradour, R.~Collobert, and J.~Weston.
\newblock Curriculum learning.
\newblock \emph{ICML}, 2009.

\bibitem[Bertinetto et~al.(2016)Bertinetto, Henriques, Valmadre, Torr, and
  Vedaldi]{learnet}
L.~Bertinetto, J.~Henriques, J.~Valmadre, P.~Torr, and A.~Vedaldi.
\newblock Learning feed-forward one-shot learners.
\newblock \emph{NIPS}, 2016.

\bibitem[Carlson et~al.(2010)Carlson, Betteridge, Kisiel, Settles, Hruschka,
  and Mitchell]{nell}
A.~Carlson, J.~Betteridge, B.~Kisiel, B.~Settles, E.~R. Hruschka, and T.~M.
  Mitchell.
\newblock Toward an architecture for never-ending language learning.
\newblock \emph{AAAI}, 2010.

\bibitem[Caruana(1998)]{multitask}
R.~Caruana.
\newblock Multitask learning.
\newblock In \emph{Learning to learn}. Springer, 1998.

\bibitem[Denoyer and Gallinari(2015)]{denoyer15}
L.~Denoyer and P.~Gallinari.
\newblock Deep sequential neural networks.
\newblock \emph{EWRL}, 2015.

\bibitem[Dorn(1960)]{dorn1960duality}
W.~S. Dorn.
\newblock Duality in quadratic programming.
\newblock \emph{Quarterly of Applied Mathematics}, 1960.

\bibitem[Eigen et~al.(2014)Eigen, Sutskever, and Ranzato]{eigen14}
D.~Eigen, I.~Sutskever, and M.~Ranzato.
\newblock Learning factored representations in a deep mixture of experts.
\newblock \emph{ICLR}, 2014.

\bibitem[Fei-Fei et~al.(2003)Fei-Fei, Fergus, and Perona]{one_shot}
L.~Fei-Fei, R.~Fergus, and P.~Perona.
\newblock A {Bayesian} approach to unsupervised one-shot learning of object
  categories.
\newblock \emph{ICCV}, 2003.

\bibitem[{Fernando} et~al.(2017){Fernando}, {Banarse}, {Blundell}, {Zwols},
  {Ha}, {Rusu}, {Pritzel}, and {Wierstra}]{pathnet}
C.~{Fernando}, D.~{Banarse}, C.~{Blundell}, Y.~{Zwols}, D.~{Ha}, A.~A. {Rusu},
  A.~{Pritzel}, and D.~{Wierstra}.
\newblock {PathNet}: Evolution channels gradient descent in super neural
  networks.
\newblock \emph{arXiv}, 2017.

\bibitem[French(1999)]{catastrophic_french}
R.~M. French.
\newblock Catastrophic forgetting in connectionist networks.
\newblock \emph{Trends in cognitive sciences}, 1999.

\bibitem[{Goodfellow} et~al.(2013){Goodfellow}, {Mirza}, {Xiao}, {Courville},
  and {Bengio}]{catastrophic_empirical}
I.~J. {Goodfellow}, M.~{Mirza}, D.~{Xiao}, A.~{Courville}, and Y.~{Bengio}.
\newblock {An Empirical Investigation of Catastrophic Forgetting in
  Gradient-Based Neural Networks}.
\newblock \emph{arXiv}, 2013.

\bibitem[He et~al.(2015)He, Zhang, Ren, and Sun]{resnet}
K.~He, X.~Zhang, S.~Ren, and J.~Sun.
\newblock Deep residual learning for image recognition.
\newblock \emph{arXiv}, 2015.

\bibitem[Hinton et~al.(2015)Hinton, Vinyals, and Dean]{distillation}
G.~Hinton, O.~Vinyals, and J.~Dean.
\newblock Distilling the knowledge in a neural network.
\newblock \emph{arXiv}, 2015.

\bibitem[{Jung} et~al.(2016){Jung}, {Ju}, {Jung}, and {Kim}]{less_forgetting}
H.~{Jung}, J.~{Ju}, M.~{Jung}, and J.~{Kim}.
\newblock {Less-forgetting Learning in Deep Neural Networks}.
\newblock \emph{arXiv}, 2016.

\bibitem[Kirkpatrick et~al.(2017)Kirkpatrick, Pascanu, Rabinowitz, Veness,
  Desjardins, Rusu, Milan, Quan, Ramalho, Grabska-Barwinska, et~al.]{ewc}
J.~Kirkpatrick, R.~Pascanu, N.~Rabinowitz, J.~Veness, G.~Desjardins, A.~A.
  Rusu, K.~Milan, J.~Quan, T.~Ramalho, A.~Grabska-Barwinska, et~al.
\newblock Overcoming catastrophic forgetting in neural networks.
\newblock \emph{PNAS}, 2017.

\bibitem[Krizhevsky(2009)]{cifar100}
A.~Krizhevsky.
\newblock Learning multiple layers of features from tiny images.
\newblock Technical report, Technical report, University of Toronto, 2009.

\bibitem[Lampert et~al.(2009)Lampert, Nickisch, and Harmeling]{zero_lampert}
C.~Lampert, H.~Nickisch, and S.~Harmeling.
\newblock Learning to detect unseen object classes by between-class attribute
  transfer.
\newblock \emph{CVPR}, 2009.

\bibitem[LeCun et~al.(1998)LeCun, Cortes, and Burges]{mnist}
Y.~LeCun, C.~Cortes, and C.~J. Burges.
\newblock The {MNIST} database of handwritten digits, 1998.
\newblock URL \url{http://yann.lecun.com/exdb/mnist/}.

\bibitem[Li and Hoiem(2016)]{lwc}
Z.~Li and D.~Hoiem.
\newblock Learning without forgetting.
\newblock \emph{ECCV}, 2016.

\bibitem[{Lucic} et~al.(2017){Lucic}, {Faulkner }, {Krause}, and
  {Feldman}]{coresetMoG}
M.~{Lucic}, M.~{Faulkner }, A.~{Krause}, and D.~{Feldman}.
\newblock {Training Mixture Models at Scale via Coresets}.
\newblock \emph{arXiv}, 2017.

\bibitem[McClelland et~al.(1995)McClelland, McNaughton, and
  O'reilly]{hippocampus}
J.~L. McClelland, B.~L. McNaughton, and R.~C. O'reilly.
\newblock Why there are complementary learning systems in the hippocampus and
  neocortex: insights from the successes and failures of connectionist models
  of learning and memory.
\newblock \emph{Psychological review}, 1995.

\bibitem[McCloskey and Cohen(1989)]{catastrophic}
M.~McCloskey and N.~J. Cohen.
\newblock Catastrophic interference in connectionist networks: The sequential
  learning problem.
\newblock \emph{Psychology of learning and motivation}, 1989.

\bibitem[Mikolov et~al.(2015)Mikolov, Joulin, and Baroni]{commai}
T.~Mikolov, A.~Joulin, and M.~Baroni.
\newblock A roadmap towards machine intelligence.
\newblock \emph{arXiv}, 2015.

\bibitem[Oquab et~al.(2014)Oquab, Bottou, Laptev, and Sivic]{oquab}
M.~Oquab, L.~Bottou, I.~Laptev, and J.~Sivic.
\newblock Learning and transferring mid-level image representations using
  convolutional neural networks.
\newblock \emph{CVPR}, 2014.

\bibitem[Palatucci et~al.(2009)Palatucci, Pomerleau, Hinton, and
  Mitchell]{zero_palatucci}
M.~Palatucci, D.~A. Pomerleau, G.~E. Hinton, and T.~Mitchell.
\newblock Zero-shot learning with semantic output codes.
\newblock \emph{NIPS}, 2009.

\bibitem[Pan and Yang(2010)]{transfer_learning}
S.~J. Pan and Q.~Yang.
\newblock A survey on transfer learning.
\newblock \emph{TKDE}, 2010.

\bibitem[Pentina and Urner(2016)]{pentina16}
A.~Pentina and R.~Urner.
\newblock Lifelong learning with weighted majority votes.
\newblock \emph{NIPS}, 2016.

\bibitem[Pentina et~al.(2015)Pentina, Sharmanska, and
  Lampert]{curriculum_tasks}
A.~Pentina, V.~Sharmanska, and C.~H. Lampert.
\newblock Curriculum learning of multiple tasks.
\newblock \emph{CVPR}, 2015.

\bibitem[Peters et~al.(2016)Peters, B{\"u}hlmann, and Meinshausen]{icp}
J.~Peters, P.~B{\"u}hlmann, and N.~Meinshausen.
\newblock Causal inference by using invariant prediction: identification and
  confidence intervals.
\newblock \emph{Journal of the Royal Statistical Society}, 2016.

\bibitem[{Rannen Triki} et~al.(2017){Rannen Triki}, {Aljundi}, {Blaschko}, and
  {Tuytelaars}]{encoder_lifelong}
A.~{Rannen Triki}, R.~{Aljundi}, M.~B. {Blaschko}, and T.~{Tuytelaars}.
\newblock {Encoder Based Lifelong Learning}.
\newblock \emph{arXiv}, 2017.

\bibitem[Ratcliff(1990)]{connectionist}
R.~Ratcliff.
\newblock Connectionist models of recognition memory: Constraints imposed by
  learning and forgetting functions.
\newblock \emph{Psychological review}, 1990.

\bibitem[{Rebuffi} et~al.(2017){Rebuffi}, {Kolesnikov}, {Sperl}, and
  {Lampert}]{icarl}
S.-A. {Rebuffi}, A.~{Kolesnikov}, G.~{Sperl}, and C.~H. {Lampert}.
\newblock {iCaRL}: Incremental classifier and representation learning.
\newblock \emph{CVPR}, 2017.

\bibitem[Ring(1994)]{ring1994}
M.~B. Ring.
\newblock \emph{Continual Learning in Reinforcement Environments}.
\newblock PhD thesis, University of Texas at Austin, Austin, Texas 78712, 1994.

\bibitem[Ring(1997)]{child}
M.~B. Ring.
\newblock {CHILD}: A first step towards continual learning.
\newblock \emph{Machine Learning}, 1997.

\bibitem[Rusu et~al.(2016)Rusu, Rabinowitz, Desjardins, Soyer, Kirkpatrick,
  Kavukcuoglu, Pascanu, and Hadsell]{progressive}
A.~A. Rusu, N.~C. Rabinowitz, G.~Desjardins, H.~Soyer, J.~Kirkpatrick,
  K.~Kavukcuoglu, R.~Pascanu, and R.~Hadsell.
\newblock Progressive neural networks.
\newblock \emph{NIPS}, 2016.

\bibitem[Ruvolo and Eaton(2013)]{ella}
P.~Ruvolo and E.~Eaton.
\newblock {ELLA: An Efficient Lifelong Learning Algorithm}.
\newblock \emph{ICML}, 2013.

\bibitem[Santoro et~al.(2016)Santoro, Bartunov, Botvinick, Wierstra, and
  Lillicrap]{santoro16}
A.~Santoro, S.~Bartunov, M.~Botvinick, D.~Wierstra, and T.~Lillicrap.
\newblock One-shot learning with memory-augmented neural networks.
\newblock \emph{arXiv}, 2016.

\bibitem[Schaul et~al.(2015)Schaul, Horgan, Gregor, and Silver]{uvfa}
T.~Schaul, D.~Horgan, K.~Gregor, and D.~Silver.
\newblock Universal value function approximators.
\newblock \emph{ICML}, 2015.

\bibitem[Scholkopf and Smola(2001)]{scholkopf2001learning}
B.~Scholkopf and A.~J. Smola.
\newblock \emph{Learning with kernels: support vector machines, regularization,
  optimization, and beyond}.
\newblock MIT press, 2001.

\bibitem[Sch{\"o}lkopf et~al.(2016)Sch{\"o}lkopf, Janzing, and
  Lopez-Paz]{causal_compression}
B.~Sch{\"o}lkopf, D.~Janzing, and D.~Lopez-Paz.
\newblock Causal and statistical learning.
\newblock In \emph{Learning Theory and Approximation}. Oberwolfach Research
  Institute for Mathematics, 2016.

\bibitem[Sutton et~al.(2011)Sutton, Modayil, Delp, Degris, Pilarski, White, and
  Precup]{horde}
R.~S. Sutton, J.~Modayil, M.~Delp, T.~Degris, P.~M. Pilarski, A.~White, and
  D.~Precup.
\newblock Horde: A scalable real-time architecture for learning knowledge from
  unsupervised sensorimotor interaction.
\newblock \emph{The 10th International Conference on Autonomous Agents and
  Multiagent Systems}, 2011.

\bibitem[Thrun(1994)]{thrun-iros94}
S.~Thrun.
\newblock A lifelong learning perspective for mobile robot control.
\newblock \emph{Proceedings of the IEEE/RSJ/GI Conference on Intelligent Robots
  and Systems}, 1994.

\bibitem[Thrun(1996)]{nthing}
S.~Thrun.
\newblock Is learning the n-th thing any easier than learning the first?
\newblock \emph{NIPS}, 1996.

\bibitem[Thrun(1998)]{lifelong}
S.~Thrun.
\newblock Lifelong learning algorithms.
\newblock In \emph{Learning to learn}. Springer, 1998.

\bibitem[Thrun and Pratt(2012)]{learning_to_learn}
S.~Thrun and L.~Pratt.
\newblock \emph{Learning to learn}.
\newblock Springer Science \& Business Media, 2012.

\bibitem[Vapnik(1998)]{Vapnik98}
V.~Vapnik.
\newblock \emph{Statistical learning theory}.
\newblock Wiley New York, 1998.

\bibitem[Vinyals et~al.(2016)Vinyals, Blundell, Lillicrap, and
  Wierstra]{matchingnet}
O.~Vinyals, C.~Blundell, T.~Lillicrap, and D.~Wierstra.
\newblock Matching networks for one shot learning.
\newblock \emph{NIPS}, 2016.

\bibitem[{Zenke} et~al.(2017){Zenke}, {Poole}, and {Ganguli}]{synaptic}
F.~{Zenke}, B.~{Poole}, and S.~{Ganguli}.
\newblock {Improved multitask learning through synaptic intelligence}.
\newblock \emph{arXiv}, 2017.

\end{thebibliography}
